%% file: main.tex
\RequirePackage[svgnames]{xcolor}
\documentclass[11pt,letterpaper]{mystyle}
\input{macro}
\input{commands}

\title{\BenchName{}: Trimming the Long-Tail of \\ Visual World Modeling Evaluation}
\runningtitle{Trimming the Long-Tail of Visual World Modeling Evaluation}

\input{authors}

\addtocontents{toc}{\protect\setcounter{tocdepth}{-1}}

\begin{document} 
\maketitle

\begin{figure}[h]
\centering
\includegraphics[width=\linewidth]{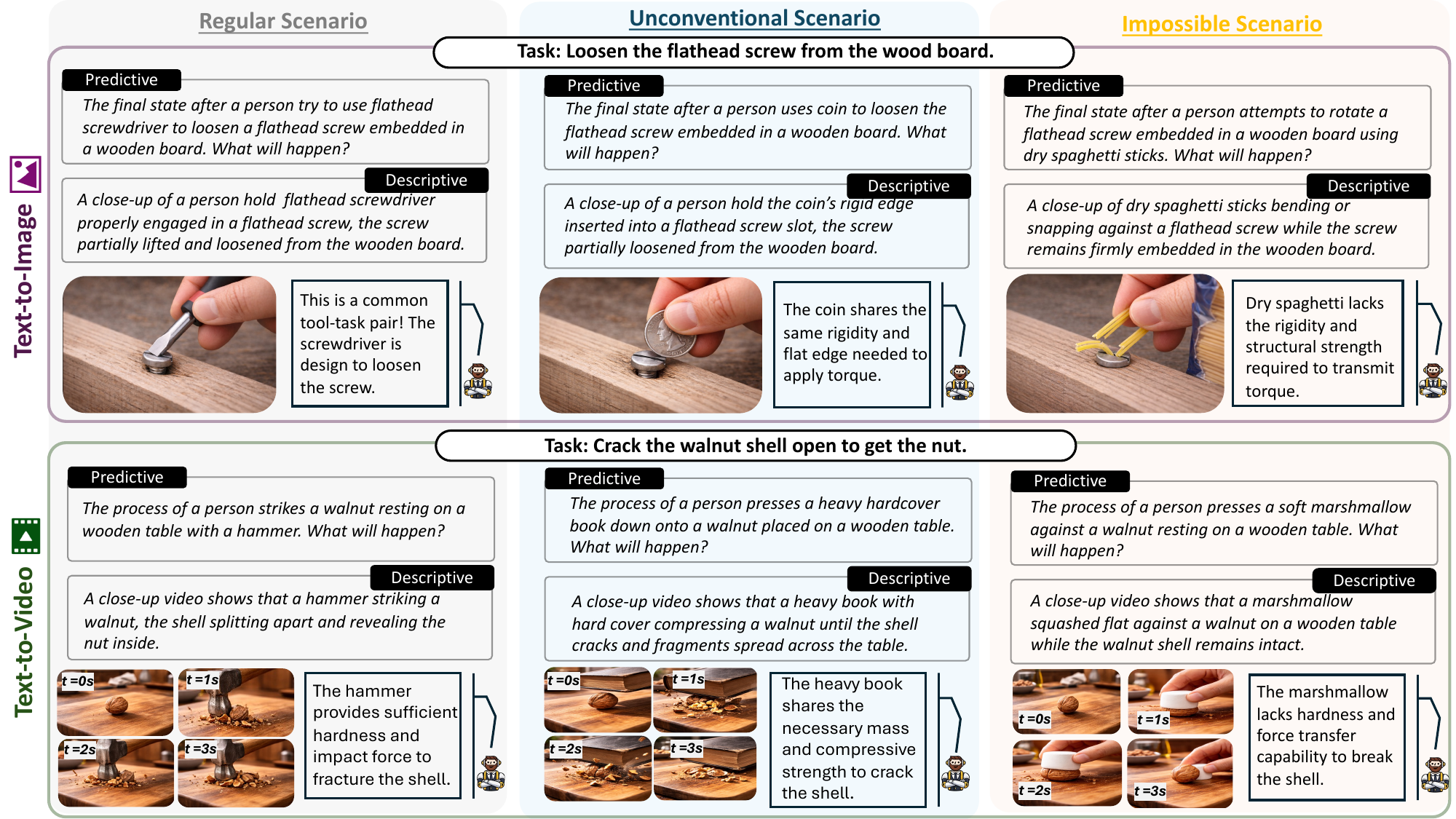}
\caption{\BenchName{} challenges world models (e.g. image generation models, video generation models) to simulate \textit{long-tail} scenarios (i.e irregular physical interactions) that require reasoning about object attributes and physical affordances beyond the typical training data distribution. Our benchmark aims to investigate: 
\textbf{\textit{Do world models truly internalize the underlying physical principles of object interactions, or do they primarily rely on statistical regularities observed in training data?}}}
\label{fig:teaser}
\vspace{-0.1in}
\end{figure}

\section*{Abstract}
\input{content/00-abstract}    

\input{content/01-introduction}
\input{content/02-related_work}
\input{content/03-benchmark}
\input{content/04-evaluation}
\input{content/05-discussion}
\input{content/06-conclusion}

\clearpage
\bibliography{main}

\clearpage
\appendix
\addtocontents{toc}{\protect\setcounter{tocdepth}{3}}
\phantomsection
\hypertarget{toc}{}
\tableofcontents

\pagestyle{fancy}
\renewcommand{\headrulewidth}{0pt}
\fancyhead{}
\fancyfoot[L]{\hyperlink{toc}{Back to Table of Contents}}
\fancyfoot[R]{\hyperlink{abstract}{Back to the First Page}}

\clearpage
\input{content/supp}


\end{document}

%% file: macro.tex
\usepackage[all]{hypcap}
\usepackage[svgnames]{xcolor}
\usepackage[comma,authoryear,compress]{natbib}
\usepackage{hyperref}[citecolor=lightblue]

\hypersetup{
    colorlinks = true,
    citecolor = {YaleBlue},
}

\usepackage{algorithm}
\usepackage{algorithmicx}
\usepackage{algpseudocode}
\usepackage{microtype}
\usepackage{graphicx}
\expandafter\def\csname ver@subfig.sty\endcsname{}
\usepackage{booktabs} %
\usepackage{float}
\usepackage{bigstrut}

\usepackage{amsmath}
\usepackage{amssymb}
\usepackage{mathtools}
\usepackage{amsthm}
\usepackage{mathrsfs}
\usepackage{nicefrac}
\usepackage{dsfont}
\usepackage{subcaption}
\usepackage{graphicx,subfig}
\usepackage{cleveref}
\usepackage{bxcoloremoji}
\usepackage{listings}
\usepackage{tcolorbox}
\tcbuselibrary{listings}

\newtcblisting{promptbox}[2][]{
    listing only,
    listing options={
        basicstyle=\ttfamily\footnotesize,
        breaklines=true,
    },
    colback=gray!10,
    colframe=YaleBlue!60, 
    title=#2,
    fonttitle=\bfseries,
    #1
}
\usepackage{enumitem}
\setlist[itemize]{leftmargin=*}
\setlist[enumerate]{leftmargin=*}
\usepackage[normalem]{ulem}

\usepackage{float}
\usepackage{multirow}
\usepackage{afterpage}

\usepackage[utf8]{inputenc} %
\usepackage[T1]{fontenc}    %
\usepackage{hyperref}       %
\usepackage{url}            %
\usepackage{booktabs}       %
\usepackage{amsfonts}       %
\usepackage{nicefrac}       %
\usepackage{microtype}      %
\usepackage{graphicx}
\usepackage{subcaption} 
\usepackage{amssymb}
\usepackage{fdsymbol}
\usepackage{wrapfig}
\usepackage{lipsum}
\usepackage{stackengine}
\usepackage[font=small,labelfont=bf]{caption}
\usepackage{color}
\usepackage{adjustbox}

\usepackage{rotating}
\usepackage{makecell}
\usepackage{listings}  
\newcommand{\x}{\mathbf{x}}

\definecolor{blanchedalmond}{rgb}{1.0, 0.92, 0.8}
\definecolor{carmine}{rgb}{0.59, 0.0, 0.09}
\definecolor{lightblue}{rgb}{0.22,0.45,0.70}%

\renewcommand{\mathbf}{\boldsymbol}

\makeatletter
\def\Ddots{\mathinner{\mkern1mu\raise\p@
\vbox{\kern7\p@\hbox{.}}\mkern2mu
\raise4\p@\hbox{.}\mkern2mu\raise7\p@\hbox{.}\mkern1mu}}
\makeatother

\definecolor{amaranth}{rgb}{0.9, 0.17, 0.31}
\definecolor{antiquebrass}{rgb}{0.8, 0.58, 0.46}
\definecolor{antiquefuchsia}{rgb}{0.57, 0.36, 0.51}
\definecolor{chromeyellow}{rgb}{0.31, 0.47, 0.26}

\newtcolorbox{AIbox}[2][]{aibox,title=#2,#1}
\definecolor{lightblue}{rgb}{0.22,0.45,0.70}%
\definecolor{Gray}{gray}{0.95}
\definecolor{Cornsilk}{rgb}{1.0, 0.97, 0.86}

\usepackage{amsmath}

\usepackage{pdfrender}

\definecolor{Gray}{gray}{0.95}
\newcolumntype{a}{>{\columncolor{Gray}}c}

\usepackage{skak}
\usepackage{bbm}
\usepackage{lipsum}
\usepackage{listings}
\usepackage{caption}
\usepackage{fancyhdr}

\definecolor{pink}{HTML}{fc6c85}

\definecolor{customblue}{HTML}{286dc0}

\definecolor{customred}{HTML}{CC0000}

\definecolor{customyellow}{HTML}{ffd55a}

\definecolor{customgrey}{HTML}{978d85}

\newtcolorbox{blueBox}[1][]{
  colback=YaleBlue!12!white,  
  colframe=YaleBlue!85,   
  boxrule=1pt,
  arc=3pt,
  floatplacement=floating,
  title=\centering #1
}

\newtcolorbox{yellowBox}[1][]{
  colback=customyellow!10!white,
  colframe=customyellow,
  floatplacement=floating,
  title=\centering #1
}

\newtcolorbox{promptBox}[1][]{
  colback=YaleBlue!12!white,  
  colframe=YaleBlue!85,      
  boxrule=0.8pt,
  arc=3pt,
  floatplacement=floating,
  title=\centering #1
}

\newtcolorbox{wronganswer}[1][]{
    enhanced,
    breakable,
    colframe=customred!65,
    colback=customred!6!white,
    sharp corners,
    boxsep=0pt,
    left=5pt,
    right=5pt,
    top=6pt,
    bottom=6pt,
    boxrule=0pt,
    leftrule=4pt,
    #1
}

\newtcolorbox{correctanswer}[1][]{
    enhanced,
    breakable,
    colframe=OliveGreen,
    colback=OliveGreen!10!white,
    sharp corners,
    boxsep=0pt,
    left=5pt,
    right=5pt,
    top=6pt,
    bottom=6pt,
    boxrule=0pt,
    leftrule=4pt,
    #1
}

\usepackage{fontawesome}

%% file: commands.tex
\usepackage{times}
\usepackage{latexsym}
\usepackage[T1]{fontenc}
\usepackage[utf8]{inputenc}
\usepackage{microtype}
\usepackage{inconsolata}
\usepackage{graphicx}

\usepackage{multirow}
\usepackage{amsmath}
\usepackage{pdfpages}
\usepackage{afterpage}
\usepackage{stfloats}
\usepackage{amsmath}
\usepackage{xspace}
\usepackage[normalem]{ulem}
\useunder{\uline}{\ul}{}
\usepackage{enumitem}
\usepackage[table]{xcolor}
\usepackage{colortbl}
\usepackage{amssymb}
\usepackage{makecell}
\usepackage{xfp} 

\usepackage{listings} 
\lstdefinelanguage{yaml}{
  keywords={true,false,null,y,n},
  sensitive=false,
  comment=[l]{\#},
  morestring=[b]',
  morestring=[b]"
}

\usepackage{multirow}
\usepackage{caption}
\usepackage{booktabs}

\usepackage[table]{xcolor}
\definecolor{tealA}{RGB}{0,128,128}
\definecolor{navyH}{RGB}{212,175,55}
\newcommand{\Ahi}[1]{\textcolor{tealA}{\textbf{#1}}}
\newcommand{\Hhi}[1]{\textcolor{navyH}{\textbf{#1}}}

\usepackage{booktabs}
\definecolor{groupgray}{gray}{0.93}   
\definecolor{lightgray}{gray}{0.97}   

\newcommand{\BenchName}{\textcolor{darkgray}{\textbf{\textsc{TailOR}}}}

\NewDocumentCommand{\ember}
{ mO{} }{\textcolor{purple}{\textsuperscript{\textit{Ember}}\textsf{\textbf{\small[#1]}}}}
\NewDocumentCommand{\cheng}
{ mO{} }{\textcolor{blue}{\textsuperscript{\textit{Cheng}}\textsf{\textbf{\small[#1]}}}}

\NewDocumentCommand{\heng}
{ mO{} }{\textcolor{red}{\textsuperscript{\textit{Heng}}\textsf{\textbf{\small[#1]}}}}

\usepackage{listings}
\usepackage[T1]{fontenc}
\usepackage{graphicx}

%% file: authors.tex
\newcommand{\uiuc}{1}
\newcommand{\stanford}{2}
\newcommand{\cu}{3}

\author[\uiuc]{Bingxuan Li}
\author[\stanford]{Yining Hong} 
\author[\uiuc]{Cheng Qian}
\author[\uiuc]{Hyeonjeong Ha}
\author[\uiuc]{\\Jiateng Liu}
\author[\uiuc]{Zhenhailong Wang}
\author[\uiuc]{Yue Guo}
\author[\cu]{Yunzhu Li}
\author[\uiuc]{Heng Ji}

\affil[\uiuc]{University of Illinois Urbana-Champaign}
\affil[\stanford]{Stanford University}
\affil[\cu]{Columbia University}

%% file: content/00-abstract.tex
Physical interactions follow a long-tailed distribution: a set of common and regular interactions dominates human experience and visual data, while a broad spectrum of rare and irregular interactions remains underrepresented. Although recent world models achieve impressive realism on existing benchmarks, they primarily focus on simulating common physical interactions. This raises a central question: \textit{Do current world models internalize the underlying physical principles of object interactions, or do they primarily rely on statistical regularities observed in training data?}
In this work, we introduce \BenchName{}, a benchmark that challenges world models to simulate irregular physical interactions. To enable systematic evaluation, we design three scenario modes that progressively challenge their reasoning: Regular scenarios reflect common tool–task pairs, unconventional scenarios replace regular tools with attribute-compatible substitutes to test affordance generalization, and impossible scenarios introduce attribute-violating tools to probe constraint awareness. Additionally, we design two complementary settings under a unified evaluation protocol: predictive generation requires inferring outcomes without guidance, while descriptive generation specifies the target outcome for faithful realization. Our experimental results reveal a clear long-tail gap in physical world modeling: Performance consistently degrades from Regular to Unconventional and Impossible scenarios, indicating limited generalization beyond common interactions. The largest drops occur in Interaction Accuracy and Physical Realism, pointing to weak affordance-level reasoning rather than perceptual issues. Failure analysis further shows that models rely on superficial visual patterns, where image models fail to realize correct state changes, and video models additionally suffer from temporal inconsistencies. These limitations persist even with explicit outcome descriptions: Models often revert to familiar patterns in descriptive settings and fail to infer correct outcomes in predictive settings. By exposing systematic failures under long-tail scenarios, \BenchName{} highlights the need for models that understand object attributes, causal dynamics, and constraint-aware interactions, rather than relying on statistical co-occurrence.

%% file: content/01-introduction.tex
\section{Introduction}
\label{sec:intro}

Physical interactions in the real world do not occur uniformly. Instead, they follow a \textit{long-tailed distribution}: A set of common interactions, such as cutting bread with a knife, a ball shattering glass under gravity, hammering a nail with a hammer, dominates both human experience and large-scale visual data. These familiar patterns form the head of the distribution, and we refer to these cases as \textbf{head scenarios}. Beyond them lies a vast and diverse \emph{long tail}: physical interactions that are less typical, improvised, or context-dependent, yet physically valid. We refer to these cases as \textbf{long-tail scenarios}. Tool-using interactions in everyday environments provide concrete illustrative examples. As illustrated in Figure\ref{fig:teaser}, a coin can substitute for a screwdriver if its rigid edge transfers sufficient torque, and a heavy hardcover book can function as a hammer to crack a walnut. These cases are rare in isolation, but collectively they occupy a large portion of the possible physical interaction space in the real world. 


Recent world models, such as image and video generation models, have advanced rapidly in producing realistic scenes to simulate real-world physical interactions~\citep{deepmind2025veo3, openai2025sora2, qwen2024image, openaiImageGeneration}. 
Current evaluations on these models largely operate under standard world assumptions (i.e., the head scenarios)~\citep{bansal2024videophy, gu2025phyworldbench, cai2025mmgr}: objects behave in familiar ways, tools are used for their intended purposes, and physical outcomes conform to everyday experience. Under such head-distribution scenarios, current models perform impressively. For example, on MMGR~\citep{cai2025mmgr} dataset, Sora-2~\citep{openai2025sora2} achieves 86\% physical accuracy and 92\% visual realism, suggesting strong capability in simulating common physical interactions. \textit{But what if the object behaves in unfamiliar ways? What happens when objects interact in unconventional ways? Can visual generation models simulate these long-tailed scenarios?} Success in head scenarios is often indistinguishable from genuine mastery of physical principles because performance in common settings can be achieved through pattern recognition and co-occurrence matching. 
Long-tail scenarios probe a fundamentally different capability. Rather than testing whether a model recalls frequent physics patterns, they require reasoning about \textit{why} an interaction works, which is largely overlooked in current evaluation benchmarks. This motivates our central question: 


\begin{center}
\textbf{Do world models internalize and generalize the physical principles?}
\end{center}

To answer this question, we introduce \BenchName{}, a benchmark designed to evaluate world models under irregular physical interactions. Our framework consists of three key components. First, we carefully define the problem and propose a \textbf{multi-dimensional benchmark design} that isolates key evaluation axes. We construct scenarios that progressively depart from familiar interactions. We begin with \textit{Regular} scenarios that mirror common tool–task pairs frequently seen in visual data. We then move to \textit{Unconventional} scenarios, where canonical tools are replaced with attribute-compatible substitutes, forcing models to go beyond surface-level associations and rely on object properties. Finally, \textit{Impossible} scenarios use attribute-violating tools to test whether models can respect physical constraints and recognize when an interaction should fail. To further understand how world models infer physical outcomes versus realize specified results, we evaluate each scenario under two complementary settings: \textit{Predictive generation}, which withholds the outcome to test outcome prediction, and \textit{Descriptive generation}, which specifies the desired result to test controllability and physical consistency.
Second, we develop a \textbf{scalable data generation pipeline} to construct diverse evaluation instances. Starting from action-driven tool-use tasks, we generate unconventional tools that satisfy required attributes and impossible tools that violate them, enabling controlled long-tail scenarios.
Third, we design a \textbf{diagnostic evaluation protocol} with rubric-based questions measuring instruction adherence, interaction correctness, physical realism, and perceptual quality. We implement an automatic evaluation pipeline using a vision-language-model as a judge and verify that its scores strongly align with human annotations. Together, these components provide a controlled testbed for diagnosing the physical principle knowledge and physical reasoning capabilities of current world models. 

\vspace{-0.02in}

We systematically evaluate state-of-the-art image models (Z-Image, Qwen-Image, GPT-Image-1, Nano-Banana-2) and video models (HunyuanVideo-1.5, Wan2.2, Sora-2, Veo-3.1) on \BenchName{}. Our experiments results \emph{trim down the long-tail} of physical world modeling, revealing a fundamental limitation of current visual generative models: a pronounced performance gap under long-tail interactions. Across both image and video generation, performance degrades consistently from Regular to Unconventional and Impossible scenarios, highlighting limited generalization beyond common, head-distribution patterns. The drop is most severe in Interaction Accuracy and Physical Realism, suggesting that failures stem from weak affordance-level understanding rather than perceptual quality. Failure analysis shows that models rely on superficial visual patterns instead of physically grounded reasoning. Image models often produce plausible scenes but fail to realize correct state changes or object attributes, while video models suffer from additional temporal failures such as implausible dynamics and cascading inconsistencies over time. These limitations persist even when the desired outcome is explicitly specified. In descriptive settings, models frequently ignore instructions and revert to familiar interaction patterns, revealing a bias toward perceptual realism over causal consistency. In contrast, failures in predictive settings indicate a lack of underlying physical reasoning, as models struggle to infer correct outcomes without guidance. Together, these findings suggest that current world models largely memorize interaction templates rather than perform compositional physical reasoning, limiting their ability to handle long-tail scenarios.


%% file: content/02-related_work.tex
\section{Related Work}
\label{sec:related_work}

\noindent\textbf{World Modeling with Multimodal Generative Models. }
Recent progress in multimodal generative models has led to significant improvements in image and video synthesis quality. Large-scale image generation models such as Qwen-Image~\citep{qwen2024image}, Gemini Image~\citep{google2025nano}, and GPT-Image-1~\citep{openaiImageGeneration} demonstrate strong capabilities in producing high-fidelity images that align with textual prompts. In the video domain, early transformer-based approaches such as CogVideo~\citep{hong2022cogvideo} laid the foundation for text-to-video generation, while recent large-scale models, including Wan~\citep{wan2025video}, MovieGen~\citep{polyak2024movie}, Kling~\citep{kling2024}, and Pika~\citep{pika2024}, further improve temporal consistency and controllability. More recent systems such as Veo~3~\citep{deepmind2025veo3}, Sora~2~\citep{openai2025sora2}, and Seedance~\citep{seedance2025seedance} extend generation to longer sequences with improved physical realism and audio-visual coherence. Beyond visual synthesis, several works explore the role of generative models as implicit world simulators. A recent study argues that large video generation models may learn internal representations resembling world simulation~\citep{videoworldsimulators2024}. Related efforts also investigate how generated videos can support downstream planning or control tasks~\citep{chen2025large}. Despite these advances, it remains unclear whether improved perceptual realism reflects genuine understanding of physical constraints and object affordances or merely improved pattern reproduction, motivating systematic evaluation of implicit world knowledge in generative models.

\noindent\textbf{Evaluation of Multimodal Generative Models. }
A growing body of work studies evaluation methodologies for text-to-image and text-to-video generation. Image generation benchmarks such as T2I-CompBench~\citep{huang2023t2i}, GenEval~\citep{ghosh2023geneval}, and GenAI-Bench~\citep{li2024genai} focus on compositionality, object-level correctness, and prompt alignment, while TIFA~\citep{hu2023tifa} introduces question-answering-based evaluation for interpretable faithfulness assessment. GECKO~\citep{wiles2024revisiting} further highlights challenges in evaluation design, including prompt sensitivity and the limitations of automatic metrics and human judgments. For video generation, recent benchmarks expand evaluation toward temporal consistency and physical reasoning. VideoPhy and VideoPhy-2~\citep{bansal2024videophy,bansal2025videophy} evaluate physical commonsense reasoning in generated videos, while PhyWorldBench~\citep{gu2025phyworldbench} and Morpheus~\citep{zhang2025morpheus} examine physical realism through structured scenarios and real-world experiments. Other benchmarks, including VBench-2.0~\citep{zheng2025vbench}, Video-Bench~\citep{han2025video}, and UI2V-Bench~\citep{zhang2025ui2v}, focus on intrinsic faithfulness, human alignment, and understanding-based generation quality. Recent studies further investigate whether video models exhibit reasoning or zero-shot generalization capabilities~\citep{wiedemer2025video,guo2025mmecof}.  MMGR~\citep{cai2025mmgr} explores multimodal generative reasoning more broadly. While these efforts significantly advance evaluation, most existing benchmarks emphasize perceptual quality, prompt faithfulness, or physical correctness in conventional scenarios. In contrast, our work additionally evaluates unconventional yet physically plausible settings that require models to generalize object affordances and implicit world constraints beyond common visual distributions.


%% file: content/03-benchmark.tex
\section{Problem Formulation}
\label{sec:problem}

Recent world models (e.g., image and video generation models) are capable of generating visually realistic physical interaction scenes \citep{cai2025mmgr, videoworldsimulators2024, deepmind2025veo3}. However, it remains unclear whether such performance reflects genuine internalization of physical principles or reliance on statistical regularities in the training data.
We investigate this question through tool-use tasks: Regular tool–task pairings dominate visual data and constitute the head of the distribution, which we define as \emph{\textbf{head scenarios}}. In contrast, irregular tool-use cases require reasoning over object attributes and physical affordances, such as rigidity, geometry, and structural strength, and therefore lie in the \emph{long tail}. We define these irregular tool–task pairings as \emph{\textbf{long-tail scenarios}}.

Following this formulation, we curate a benchmark with long-tail scenarios that serve as a testbed to examine whether world models genuinely internalize physical principles: If a substitute object satisfies the required physical attributes to complete a certain task, generation should succeed even when the pairing is rare; Conversely, if it violates critical constraints, the outcome should accurately and consistently reflect the failure. Formally, each evaluation instance is defined as $x = (g, r, \mathcal{A}, \mathcal{U}, \mathcal{I})$, where $g$ denotes the task goal (e.g., ``hold a door open,'' ``block the sink''), $r$ is the canonical tool, $\mathcal{A}$ specifies the required functional attributes, $\mathcal{U}$ contains unconventional but attribute-compatible substitutes, and $\mathcal{I}$ contains attribute-violating and physically-incompatible tools.

\noindent\textbf{Scenario Modes.}
To systematically evaluate model behavior under different distributional conditions, we decompose long-tail scenarios into two sub-modes and define three scenario modes that progressively probe in-distribution performance, internalization of physical principles, and attribute-level generalization:

\begin{itemize}
    \item \textbf{\textit{Regular Scenario}.} 
    The regular tool $r$ is provided, representing frequent, in-distribution interactions. This setting evaluates whether the model can reproduce common tool--task patterns that are likely well represented in training data. This scenario belongs to head scenario category.

    \item \textbf{\textit{Unconventional Scenario}.} 
    A substitute tool $u \in \mathcal{U}$ satisfies the required attributes in $\mathcal{A}$ but is atypical for the task. This condition tests whether the model can generalize based on functional attributes rather than relying on memorized tool--task associations. This scenario belongs to long-tail scenario category.

    \item \textbf{\textit{Impossible Scenario}.} 
    A tool $i \in \mathcal{I}$ violates one or more critical attributes in $\mathcal{A}$. This setting evaluates whether the model respects physical constraints and generates outcomes that appropriately reflect failure or incompatibility. This scenario belongs to long-tail scenario category.
\end{itemize}

\noindent\textbf{Evaluated Model Types.}
We evaluate two types of world models: image generation models and video generation models. These two modalities reflect different levels of physical modeling complexity. Image generation models are required to produce a single, spatially coherent final state, primarily testing whether the model can represent physically plausible configurations and outcomes. Video generation models must additionally model temporal dynamics, including motion trajectories, contact events, force application, and state transitions over time.

\noindent\textbf{Evaluation Settings.}
Physical world modeling involves two complementary capabilities: (i) anticipating outcomes based on implicit physical knowledge, and (ii) faithfully realizing explicitly specified instructions. To disentangle these abilities, we design two complementary generation settings:
\begin{itemize}
    \item \textbf{\textit{Predictive Generation}.} 
    Predictive generation evaluates a model’s implicit physical reasoning by requiring it to anticipate the outcome of applying tool $X$ to object or scenario $Y$ for task $Z$, without revealing the final result. The model must rely on its internal representation of functional attributes (e.g., rigidity, sharpness, friction, structural compatibility) to determine whether the interaction should succeed or fail and what state change should occur. For example, given ``What will happen if we use a paperclip to zip up a jacket with a missing zipper pull?'', a physically grounded model should generate an outcome consistent with the tool’s affordances and the task requirements.

    \item \textbf{\textit{Descriptive Generation}.} 
    Descriptive generation evaluates controllability and physical consistency under explicitly specified outcomes. The desired result is provided in advance, and the model must generate content that faithfully realizes this target state. For image generation, we specify the required final configuration (e.g., ``Generate an image where a coin is used to tighten a flathead screw, and the screw is fully secured into the wooden surface.''). For video generation, we additionally describe key interaction dynamics to enforce temporal coherence.
\end{itemize}

The core distinction between the two evaluation settings lies in \textit{whether the expected outcome is specified}. Predictive generation requires the model to infer the outcome based solely on its internal physical understanding. In contrast, descriptive generation removes this ambiguity by explicitly providing the target result, thereby testing whether the model can adhere to stated physical constraints, maintain consistency between tool properties and scene dynamics, and produce interactions that are physically plausible given the specified outcome.

\section{Benchmark Design and Curation}
\label{sec:benchmark}

\begin{figure}[thbp]
    \centering    \includegraphics[width=1\linewidth]{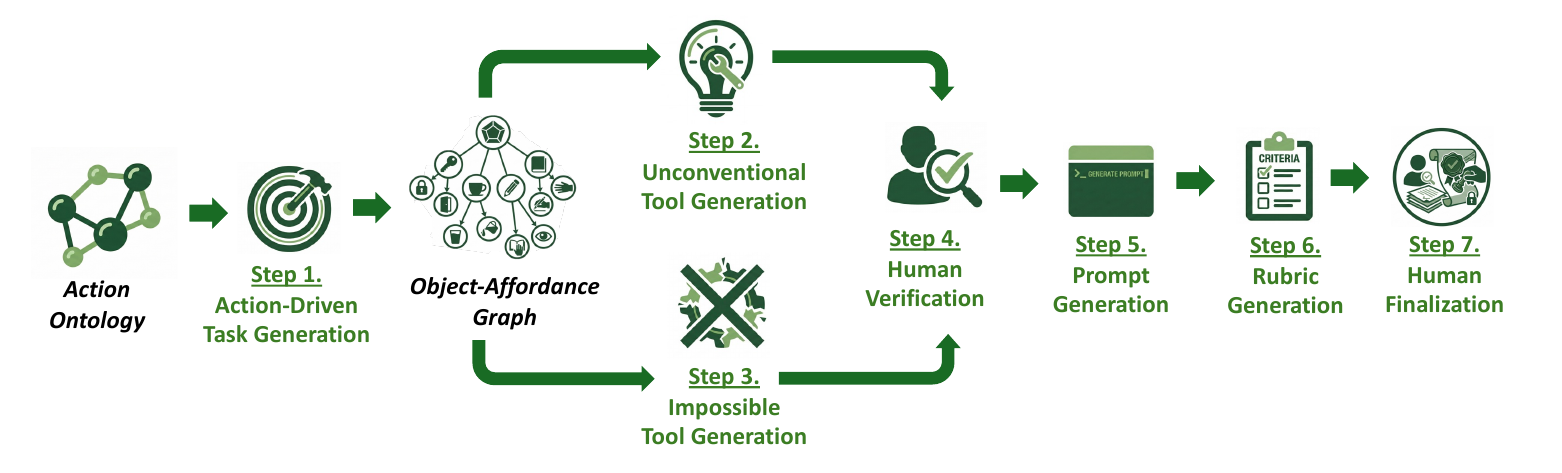}
    \caption{Data curation pipeline for \BenchName{}. From structured action and affordance resources, we construct benchmark instances via: (1) action-driven task specification; (2) LLM-generated unconventional substitutes; (3) opposite-affordance construction of impossible tools; (4) human verification and scenario balancing; (5) prompt generation (Predictive, Descriptive); (6) automatic rubric generation; and (7) final human quality control.}
    \label{fig:data_pipeline}
    \vspace{-0.2in}
\end{figure}

Figure~\ref{fig:data_pipeline} illustrates the data curation pipeline of \BenchName{}. Our objective is to construct a scalable and structured data engine that systematically generates controlled tool-use scenarios under different constraints.

\subsection{Raw Data Sources}
We construct \BenchName{} from two structured resources that provide complementary semantic and physical grounding.

\noindent\textbf{(1) Common action ontology.}
We initialize candidate tool-use actions from HICO-DET~\citep{HICO-DET}, which provides human--object interaction (HOI) verb classes and HOI categories. From these candidates, we manually verify and filter a subset of 18 actions used in our benchmark. Each action is associated with a structured description including the \emph{action name}, \emph{action description}, \emph{underlying physics}, and \emph{core affordance requirements}.

\noindent\textbf{(2) Object--affordance graph.}
We build an object--affordance resource from ConceptNet~\citep{conceptnet} by extracting object concepts and their properties (e.g., \texttt{key} \emph{HasProperty} \texttt{sharp}). We use this graph as addtional information for candidate objects during unconventional and impossible tool generation.

\subsection{Evaluation Instance Generation}
Based on the structured foundation from raw data sources, we generate each evaluation instance based on the following steps.

\noindent\textbf{Step 1: Action-driven task generation.}
For each action, we leverage LLMs to generate a diverse set of tool-use tasks that related to the action. Each task contains the \texttt{task\_goal}, \texttt{expected\_outcome}, \texttt{original\_tool}, and \texttt{required\_tool\_attributes} (i.e., the constraints needed to accomplish task). Each task is an evaluation instance for the regular scenario.

\noindent\textbf{Step 2: Unconventional tool generation.}
Given the required attributes of a task, we then leverage LLMs to propose \emph{uncommon but feasible} substitute tools that satisfy the same attributes. This step yields \texttt{unconventional\_tools} for each task, and each task-unconventional tool pair serve as an evaluation instance for the unconventional scenario.

\noindent\textbf{Step 3: Impossible tool generation.}
We then construct physically implausible tool candidates by first generating \emph{opposite affordances} (e.g., \texttt{sharp} vs.\ \texttt{blunt/round}, \texttt{long} vs.\ \texttt{short}) using LLMs. Based on these inverted properties, we further query the object–affordance graph to identify tools that satisfy such contradictory attributes, and then generate physically impossible tool candidates. We further generate the expected failure outcomes when they are applied to the task. This step yields both \texttt{impossible\_tools} and \texttt{expected\_outcome\_impossible\_tool} for each task, and each task-impossible tool pair serve as an evaluation instance for the impossible scenario.

\noindent\textbf{Step 4: Human verification and filtering.}
In this step, human annotators start with reviewing generated tasks and remove low-quality or ambiguous instances. For each retained task, annotators select and/or revise two unconventional tools and two impossible tools to ensure (i) physical plausibility for unconventional substitutions, (ii) unambiguous impossibility for negative cases, and (iii) visual clarity for downstream image/video generation. Each task is paired with five instances: one regular scenario, two unconventional scenarios, and two impossible scenarios.

\noindent\textbf{Step 5: Prompt generation.}
For each task, we generate two types of prompt to evaluate models under two setting that introduced in section \ref{sec:problem}: \emph{Predictive Generation} and \emph{Descriptive Generation}. 

\noindent\textbf{Step 6: Evaluation rubric generation.}
For each prompt, we construct a checklist-based evaluation rubric grounded in both the prompt and the task specification. The rubric decomposes evaluation into two structured dimensions: (1) \emph{Instruction Adherence}, which assesses whether the generated content instantiates required entities, attributes, and scene constraints (e.g., tool presence, correct interaction region), and (2) \emph{Interaction Accuracy}, which evaluates whether the interaction behavior and final state are consistent with the expected outcome (e.g., correct state change, failure depiction in impossible cases). These structured questions enable objective percentage-based scoring and maintain alignment between task intent and evaluation criteria. Details of metric definitions will be introduced in Section~\ref{sec:metrics}.

\noindent\textbf{Step 7: Human verification and finalization.}
We conduct a second round of human verification to ensure dataset quality. Annotators review prompts and rubrics and correct unclear, inconsistent, or low-quality items before finalization.

\begin{figure}[bthp]
    \vspace{-0.2in}
    \centering
\begin{minipage}[b]{0.45\linewidth}
    \centering
    \vspace{0pt}
    \small
    \setlength{\tabcolsep}{2.5pt}
    \renewcommand{\arraystretch}{1.25}
    \begin{tabular}{@{}l l r@{}}
        \toprule
        \textbf{\textit{Scenario}}
        & \textbf{\textit{Category}}
        & \textbf{\textit{Count}} \\
        \midrule
        \multirow{2}{*}{\textit{Regular}}
            & Evaluation Tasks & 80 \\
            & Generation Prompts & 320 \\
        \midrule
        \multirow{2}{*}{\textit{Unconventional}}
            & Evaluation Tasks & 160 \\
            & Generation Prompts & 640 \\
        \midrule
        \multirow{2}{*}{\textit{Impossible}}
            & Evaluation Tasks & 160 \\
            & Generation Prompts & 640 \\
        \midrule
        \multirow{2}{*}{\textbf{Total}}
            & Evaluation Tasks & \textbf{400} \\
            & Generation Prompts & \textbf{1600} \\
        \bottomrule
    \end{tabular}
    \vspace{0.1in}
    \captionof{table}{Dataset statistics of \BenchName{}, organized by scenario type (Regular, Unconventional, Impossible). Each evaluation task is expanded into four generation prompts, yielding 400 evaluation tasks and 1,600 evaluation instances (i.e. prompts) in total.}
    \label{tab:dataset_statistics}
\end{minipage}
    \hfill
    \begin{minipage}[b]{0.52\linewidth}
        \centering
        \vspace{0pt}
        \includegraphics[width=\linewidth]{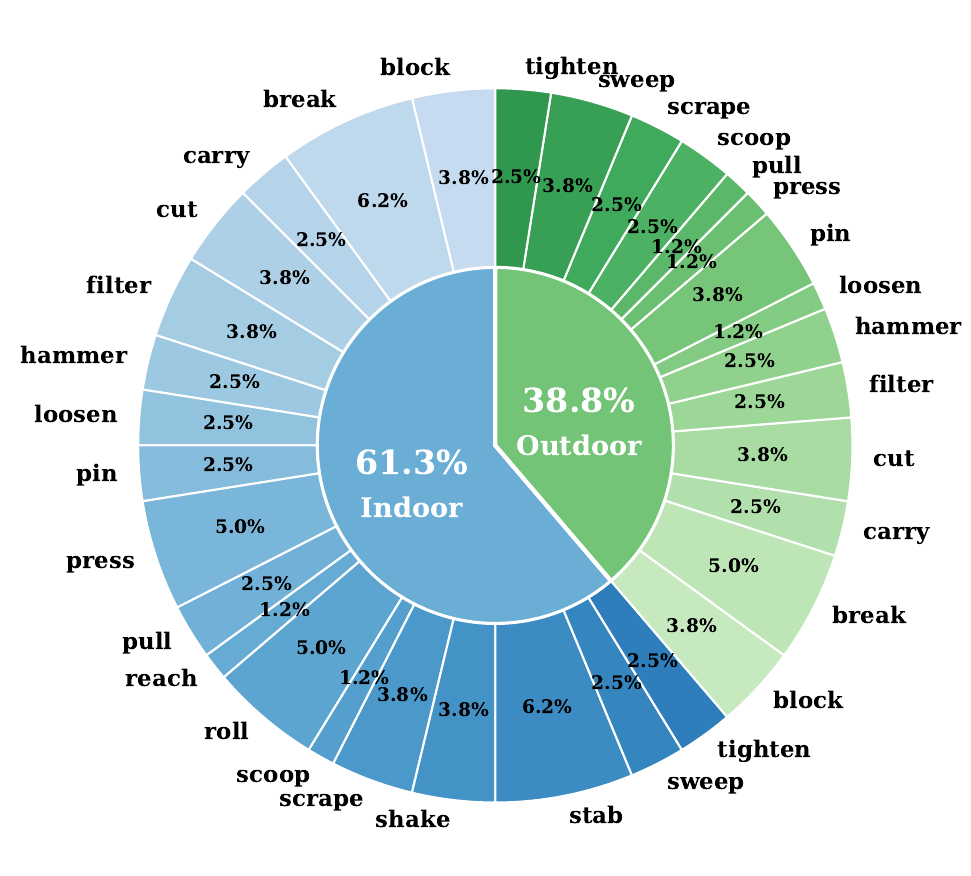}
        \caption{Action distribution in \BenchName{} across action and environment categories.}
        \label{fig:dataset_action_distribution}
    \end{minipage}
\end{figure}
\vspace{-0.2in}

\subsection{Dataset Statistics}

After two rounds of human verification and filtering, \BenchName{} contains 80 tool-use tasks spanning diverse action categories across both indoor and outdoor environments (Fig.~\ref{fig:dataset_action_distribution}). Each task is instantiated with five tool variants: One regular tool, two unconventional yet attribute-compatible substitutes, and two attribute-violating impossible tools. Each instance is evaluated for both image and video generation under two prompt types (predictive and descriptive). Thus, every task–tool pair corresponds to four generation prompts, yielding 1600 prompts in total. Table~\ref{tab:dataset_statistics} summarizes the distribution of evaluation instances and prompts across scenario types.

\subsection{Evaluation Metrics}
\label{sec:metrics}

Evaluating long-tail physical interactions requires disentangling scene construction, interaction correctness, physical consistency, and perceptual fidelity. A model may produce visually plausible content while failing to ground object affordances, realize the correct outcome, or respect causal dynamics. To diagnose these behaviors, we evaluate four complementary dimensions: (1) \textit{Instruction Adherence}, (2) \textit{Interaction Accuracy}, (3) \textit{Physical Realism}, and (4) \textit{Perceptual Quality}. We define each metric as follows:

\begin{itemize}[leftmargin=*]

    \item \textbf{Instruction Adherence (0--100\%).}  
    Measures whether the generated scene correctly instantiates the entities and functional properties required to enable the intended interaction. It is computed as the arithmetic mean of the following checklist-based sub-scores (each reported as a percentage):

    (1) \textit{\textbf{Entity Completeness}.}  
    The proportion of checklist items correctly satisfied with respect to the presence of all required entities.  
    \textit{Example checklist questions:} Is the specified tool present?

    (2) \textit{\textbf{Attribute Fidelity}.}  
    The proportion of checklist items correctly satisfied with respect to the correct instantiation of required functional attributes.  
    \textit{Example checklist questions:} Does the tool exhibit the required functional property (e.g. sharpness)? 

    (3) \textit{\textbf{Scene Validity}.}  
    The proportion of checklist items correctly satisfied with respect to spatial configuration and physically feasible arrangement.  
    \textit{Example checklist questions:} Is the relative scale between tool and object plausible?

    \item \textbf{Interaction Accuracy (0--100\%).}  
    Measures whether the interaction outcome and dynamics are correctly realized.  
    It is computed as the arithmetic mean of the following checklist-based sub-scores (each reported as a percentage):

    (1) \textit{\textbf{State Change Correctness}.}  
    The proportion of checklist items correctly satisfied with respect to the physically correct final state (or predicted outcome).  
    \textit{Example checklist question:} Does the object exhibit the correct resulting state (e.g., cracked)?

    (2) \textit{\textbf{Affordance Grounding}.}  
    The proportion of checklist items correctly satisfied with respect to whether interaction behavior aligns with object affordances.  
    \textit{Example checklist question:} Is force applied through a structurally appropriate part of the tool?

    (3) \textit{\textbf{Motion Plausibility}.}  
    The proportion of checklist items correctly satisfied with respect to temporal coherence and dynamical feasibility. This subset is only for video generation.  \textit{Example checklist question:} Is motion trajectory continuous?

    \item \textbf{Physical Realism (0--5).}  
    An open-ended rating evaluating consistency with fundamental physical principles, including gravity, contact mechanics, material constraints, and causality. A score of 0 indicates severe and obvious violations of fundamental physical laws, whereas a score of 5 indicates full consistency with real-world physics.

    \item \textbf{Perceptual Quality (0--5).}  
    An open-ended rating assessing perceptual clarity, rendering fidelity, and overall visual quality. A score of 0 indicates severely degraded visual quality, such as heavy artifacts, structural incoherence, or identity instability, whereas a score of 5 indicates clear, stable, and highly realistic rendering. For video generation, this metric additionally evaluates cross-frame identity consistency.

\end{itemize}

\subsection{Evaluation Protocol}
We design both human and automatic evaluation protocols. More details on the evaluation protocol are described in the Appendix.

\noindent\textbf{Human Evaluation.}
We recruit 9 annotators with prior experience in visual content evaluation and physical reasoning. Annotators assess each generated sample using a standardized evaluation form that includes (i) checklist-based questions for structured, percentage-based metrics, and (ii) calibrated 0--5 rating scales for open-ended quality dimensions. Detailed annotation guidelines and reference examples are provided to ensure consistency and reduce subjective drift. Each sample is independently evaluated by three annotators to improve reliability and reduce individual bias.

\noindent\textbf{Automatic Evaluation.}
In parallel, we employ \texttt{gemini-2.5-pro} as a judge model to answer the same evaluation questions used in human assessment. The judge is provided with identical instructions, checklist items, and scoring criteria, ensuring alignment between human and automatic protocols. Scores are aggregated using the same averaging procedure to facilitate direct comparison. For subjective dimensions such as \textit{Physical Realism} and \textit{Perceptual Quality}, the judge is prompted to explicitly reason before producing a final score, encouraging more grounded and structured assessments.

%% file: content/04-evaluation.tex
\section{Evaluation}
\label{sec:experiment}

\subsection{Evaluated Models}
\label{sec:models}

We evaluate a diverse set of visual generative models spanning both image and video generation. For image generation, we include Z-Image~\citep{team2025zimage}, Qwen-Image~\citep{qwen2024image},  GPT-Image-1~\citep{openai2024gpt4o}, and Nano-Banana-2~\citep{google2025nano} representing leading open and proprietary image generation models. For video generation, we evaluate HunyuanVideo-1.5~\citep{kong2024hunyuanvideo}, Wan-2.2~\citep{wan2025video}, Sora-2~\citep{openai2025sora2}, and Veo-3.1~\citep{deepmind2025veo3}, which are the current frontier of large text-to-video models.

\subsection{Evaluation Results}
\label{sec:results}

\input{tbls/main_results}

Table~\ref{tbl:main_results} reports the average scores of all evaluation metrics across evaluation instances. We have the following three observations.

\begin{figure*}[ht]
    \centering
    \includegraphics[width=0.32\textwidth]{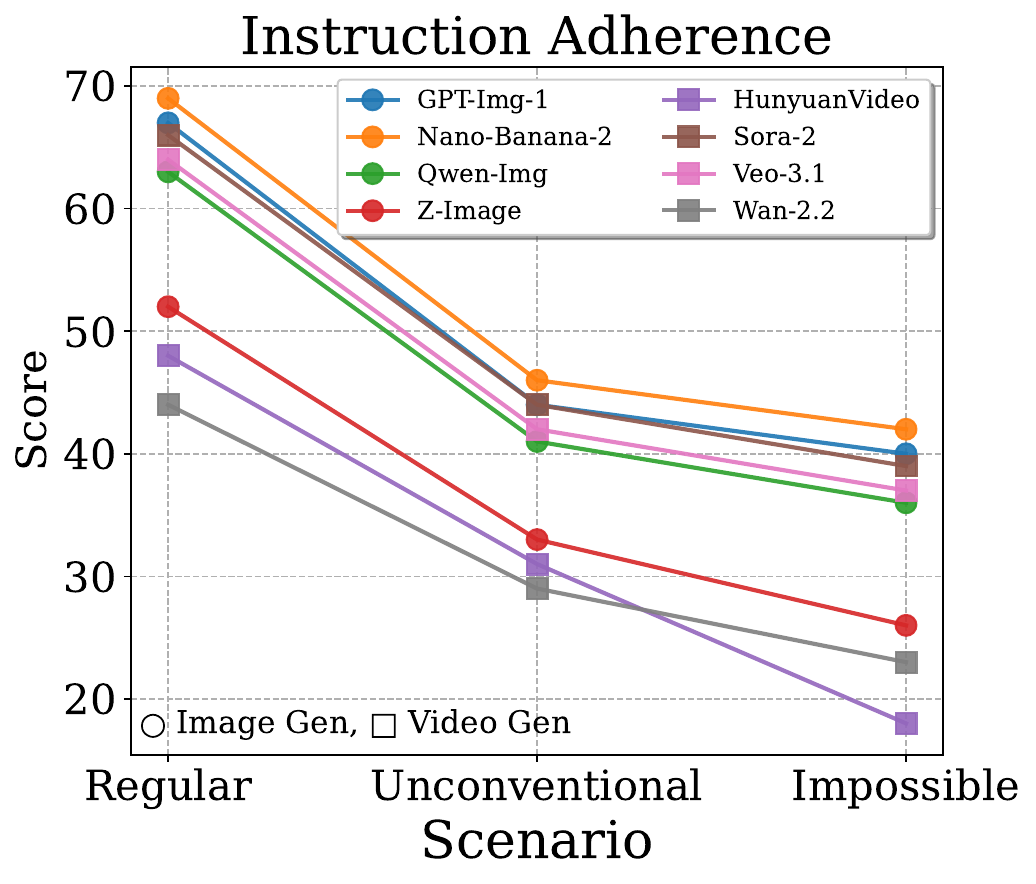}
    \includegraphics[width=0.32\textwidth]{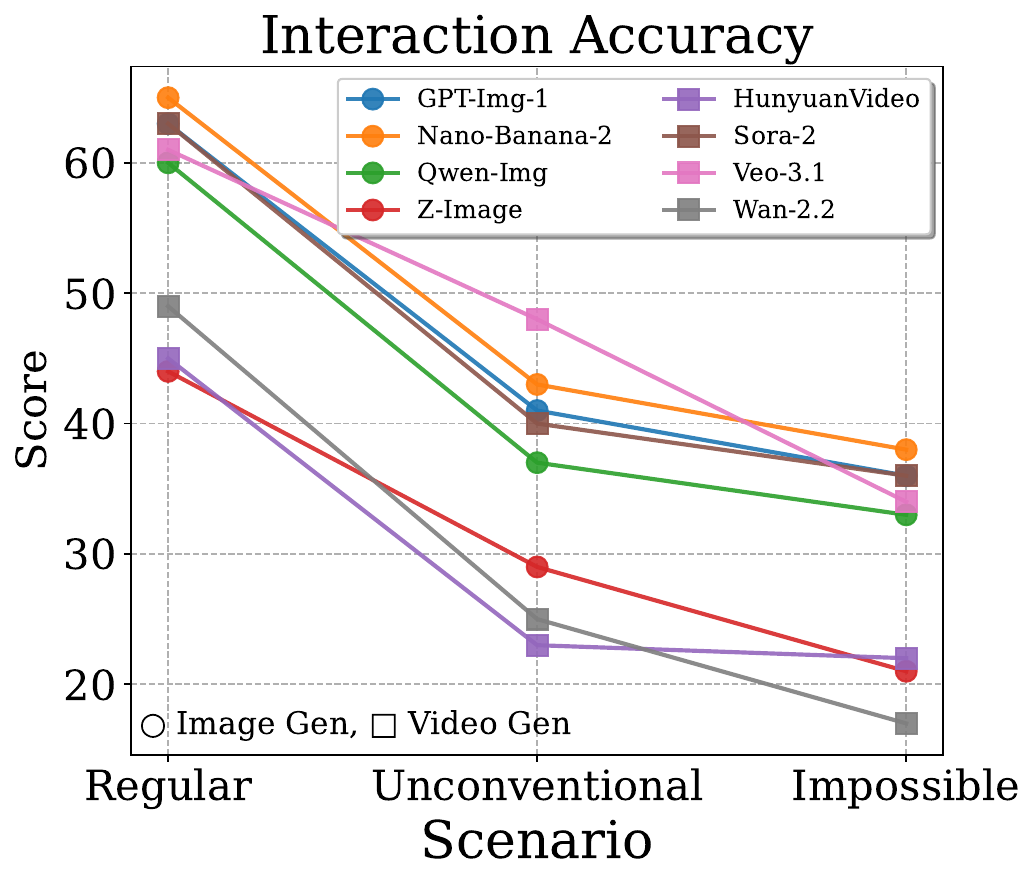}
    \includegraphics[width=0.32\textwidth]{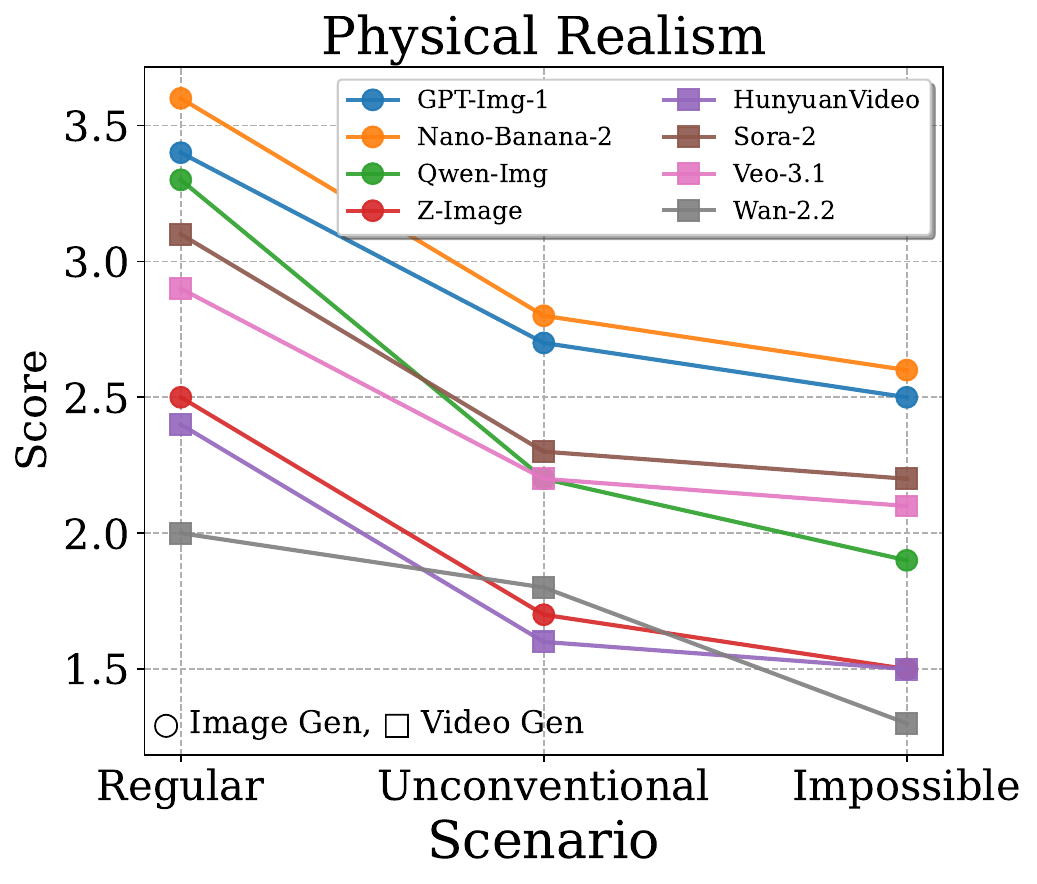}
    \caption{Predictive automatic performance across scenarios.}
    \label{fig:trends}
\end{figure*}

\noindent\textbf{\textcolor{violet}{\textit{Observation 1.} World Models Struggles at Long-Tail Scenarios.}}
Across both image and video generation models, performance consistently declines as tasks move from Regular to Unconventional and further to Impossible scenarios. As shown in Figure \ref{fig:trends}, this pattern holds across all four evaluation dimensions, indicating a systematic long-tail gap in physical world modeling. While models perform relatively better on common interactions (head-distribution), their ability to generalize to attribute-level understanding under distribution shift remains substantially limited. The degradation is especially pronounced in Interaction Accuracy and Physical Realism, suggesting that failures are not merely perceptual but stem from insufficient affordance-level understanding.

\noindent\textbf{\textcolor{violet}{\textit{Observation 2.} Greater Difficulty of Physical Reasoning in Video Models.}}
Video generation models consistently achieve lower scores than image generation models across all evaluation metrics. This gap highlights the additional difficulty of \emph{forward physical simulation} in video generation, where models must not only depict a plausible static interaction but also maintain temporally coherent state transitions across frames. Our analysis reveals that video models often exhibit a strong bias toward familiar training patterns rather than faithfully following the instruction, especially with impossible interactions. As a result, the generated dynamics frequently revert to conventional behaviors seen during training instead of executing the intended physical process. Moreover, small errors in early frames tend to propagate and amplify over time, leading to cascading failures that break causal consistency and physical plausibility. A detailed breakdown of these failure patterns is provided in Section~\ref{sec:failure_modes}.

\noindent\textbf{\textcolor{violet}{\textit{Observation 3.} Consistency Between Automatic and Human Evaluation.}}
Automatic and human evaluations exhibit strong agreement in model rankings across scenarios and metrics. As shown in Table~\ref{tbl:main_results}, most top-performing models under automatic scoring also rank among the best in human assessment, indicating that the automatic metrics reliably capture meaningful performance differences. Interestingly, both evaluation protocols consistently identify Sora-2 as the best video model and Nano-Banana-2 as the best image model across different scenarios and evaluation metrics.

%% file: tbls/main_results.tex
\begin{table}[h]
\centering
\small
\setlength{\tabcolsep}{2pt}
\renewcommand{\arraystretch}{1.45}

\resizebox{\textwidth}{!}{
\begin{tabular}{lcccccccccccc}
\toprule
\multirow{2}{*}{\textbf{Model}} 
& \multicolumn{4}{c}{\textbf{Regular (A/H)}} 
& \multicolumn{4}{c}{\textbf{Unconventional (A/H)}} 
& \multicolumn{4}{c}{\textbf{Impossible (A/H)}} \\
\cmidrule(lr){2-5} \cmidrule(lr){6-9} \cmidrule(lr){10-13}
& IA & IntAcc & Phys & Perc
& IA & IntAcc & Phys & Perc
& IA & IntAcc & Phys & Perc \\
\midrule

\rowcolor{groupgray}
\multicolumn{13}{l}{\textbf{Predictive}} \\
\midrule

\rowcolor{lightgray}
\multicolumn{13}{l}{\textit{Image Generation}} \\

\texttt{Z-Image} 
& 52\% / 55\% & 44\% / 50\% & 2.5 / 2.7 & 2.4 / 2.9
& 33\% / 39\% & 29\% / 27\% & 1.7 / 2.8 & 2.2 / 2.5
& 26\% / 24\% & 21\% / 28\% & 1.5 / 1.4 & 1.9 / 2.1 \\

\texttt{Qwen-Img}       
& 63\% / 74\% & 60\% / 70\% & 3.3 / 3.5 & 3.8 / 4.0
& 41\% / 52\% & 37\% / 48\% & 2.2 / 2.9 & 2.4 / 3.2
& 36\% / 47\% & 33\% / 43\% & 1.9 / 2.7 & 2.4 / 3.0 \\

\texttt{GPT-Img-1}      
& 67\% / 79\% & 63\% / 75\% & 3.4 / 4.2 & 3.7 / 4.4
& 44\% / 58\% & 41\% / 52\% & 2.7 / 3.3 & 2.8 / 3.6
& 40\% /  \Hhi{52\%} & 36\% / 48\% & 2.5 / 3.3 & 2.7 / 3.4 \\

\texttt{Nano-Banana-2} 
& \Ahi{69\%} / \Hhi{81\%} & \Ahi{65\%} / \Hhi{78\%} & \Ahi{3.6} / \Hhi{4.4} & \Ahi{3.8} / \Hhi{4.4}
& \Ahi{46\%} / \Hhi{60\%} & \Ahi{43\%} / \Hhi{55\%} & \Ahi{2.8} / \Hhi{3.6} & \Ahi{3.0} / \Hhi{3.7}
& \Ahi{42\%} / \Hhi{52\%} & \Ahi{38\%} / \Hhi{51\%} & \Ahi{2.6} / \Hhi{3.4} & \Ahi{2.8} / \Hhi{3.6} \\

\rowcolor{lightgray}
\multicolumn{13}{l}{\textit{Video Generation}} \\

\texttt{HunyuanVideo} 
& 48\% / 52\% & 45\% / 47\% & 2.4 / 2.3 & 2.8 / 3.2
& 31\% / 29\% & 23\% / 34\% & 1.6 / 1.8 & 2.1 / 2.6
& 18\% / 25\% & 22\% / 20\% & 1.5 / 1.6 & 1.6 / 2.4 \\

\texttt{Wan-2.2} 
& 44\% / 57\% & 49\% / 51\% & 2.0 / 2.6 & 3.0 / 2.9
& 29\% / 37\% & 25\% / 28\% & 1.8 / 2.0 & 2.5 / 2.4
& 23\% / 21\% & 17\% / 30\% & 1.3 / 1.9 & 2.2 / 2.0 \\

\texttt{Sora-2}  
& \Ahi{66\%} / \Hhi{83\%} & \Ahi{63\%} / 72\% & \Ahi{3.1} / 3.5 & \Ahi{3.4} / \Hhi{3.9}
& \Ahi{44\%} / 53\% & 40\% / \Hhi{49\%} & \Ahi{2.3} / \Hhi{2.8} & \Ahi{2.7} / \Hhi{3.2}
& \Ahi{39\%} / \Hhi{48\%} & \Ahi{36\%} / \Hhi{45\%} & \Ahi{2.2} / \Hhi{2.6} & \Ahi{2.5} / \Hhi{3.1} \\

\texttt{Veo-3.1}   
& 64\% / 75\% & 61\% / \Hhi{80\%} & 2.9 / 3.4 & 3.3 / 3.8
& 42\% / \Hhi{61\%} & \Ahi{48\%} / 47\% & 2.2 / 2.6 & 2.2 / 3.1
& 37\% / 45\% & 34\% / 42\% & 2.1 / 2.4 & 2.2 / 2.9 \\

\midrule
\rowcolor{groupgray}
\multicolumn{13}{l}{\textbf{Descriptive}} \\
\midrule

\rowcolor{lightgray}
\multicolumn{13}{l}{\textit{Image Generation}} \\

\texttt{Z-Image} 
& 54\% / 57\% & 49\% / 51\% & 2.7 / 2.6 & 2.9 / 3.1
& 30\% / 38\% & 27\% / 30\% & 1.6 / 2.0 & 2.3 / 2.2
& 21\% / 29\% & 24\% / 25\% & 1.3 / 1.7 & 1.8 / 2.0 \\

\texttt{Qwen-Img}       
& 67\% / 78\% & 63\% / 74\% & 3.1 / 3.9 & 3.3 / 4.2
& 44\% / 60\% & 41\% / 52\% & 2.3 / 3.1 & 2.6 / 3.4
& 40\% / 51\% & 36\% / 47\% & 2.1 / 2.9 & 2.4 / 3.3 \\

\texttt{GPT-Img-1}      
& \Ahi{76\%} / 85\% & 69\% / \Hhi{79\%} & 3.8 / 4.3 & 3.9 / \Hhi{4.6}
& 52\% / 62\% & 39\% / 57\% & 3.0 / 3.7 & 3.1 / 3.8
& 47\% / \Hhi{58\%} & 42\% / 52\% & 2.8 / 3.5 & 2.9 / 3.6 \\

\texttt{Nano-Banana-2} 
& 74\% / \Hhi{88\%} & \Ahi{70\%} / 76\% & \Ahi{3.9} / \Hhi{4.5} & \Ahi{4.0} / 4.5
& \Ahi{53\%} / \Hhi{65\%} & \Ahi{46\%} / \Hhi{60\%} & \Ahi{3.2} / \Hhi{4.0} & \Ahi{3.3} / \Hhi{4.2}
& \Ahi{48\%} / 56\% & \Ahi{43\%} / \Hhi{55\%} & \Ahi{3.0} / \Hhi{3.8} & \Ahi{3.2} / \Hhi{4.0} \\

\rowcolor{lightgray}
\multicolumn{13}{l}{\textit{Video Generation}} \\

\texttt{HuanyuanVideo} 
& 47\% / 53\% & 44\% / 56\% & 2.3 / 2.1 & 2.7 / 3.3
& 28\% / 31\% & 26\% / 29\% & 1.7 / 2.4 & 2.2 / 2.3
& 22\% / 24\% & 18\% / 27\% & 1.4 / 1.5 & 2.2 / 2.1 \\

\texttt{Wan-2.2} 
& 53\% / 54\% & 46\% / 58\% & 2.1 / 2.8 & 2.9 / 3.0
& 32\% / 35\% & 24\% / 36\% & 1.9 / 2.2 & 2.6 / 2.9
& 20\% / 33\% & 23\% / 26\% & 1.6 / 2.0 & 2.3 / 2.6 \\

\texttt{Sora-2}  
& \Ahi{68\%} / \Hhi{82\%} & \Ahi{64\%} / 70\% & \Ahi{3.2} / \Hhi{3.6} & \Ahi{3.6} / \Hhi{4.0}
& \Ahi{42\%} / 50\% & 38\% / \Hhi{46\%} & \Ahi{2.2} / \Hhi{2.6} & \Ahi{2.6} / \Hhi{3.1}
& \Ahi{40\%} / \Hhi{46\%} & \Ahi{37\%} / \Hhi{47\%} & \Ahi{2.3} / \Hhi{2.5} & \Ahi{2.6} / \Hhi{3.1} \\

\texttt{Veo-3}   
& 66\% / 78\% & 63\% / \Hhi{79\%} & 3.1 / 3.5 & 3.4 / 3.9
& 40\% / \Hhi{58\%} & \Ahi{46\%} / 45\% & 2.1 / 2.6 & 2.2 / 2.9
& 38\% / 44\% & 35\% / 44\% & 2.2 / 2.4 & 2.3 / 2.9 \\

\bottomrule
\end{tabular}
}
\vspace{0.1in}
\caption{ Automatic and human evaluation results across three scenarios for both prompt types under image and video generation settings. Metrics include Instruction Adherence (\textbf{IA}), Interaction Accuracy (\textbf{IntAcc}), Physical Realism (\textbf{Phys}), and Perceptual Quality (Perc). Values are reported as Automatic / Human scores (A/H). For each column under each setting, the highest automatic score is marked with \textcolor{teal}{Teal Color}, and the highest human score is marked with \textcolor{navyH}{Gold Color}. }
\label{tbl:main_results}
\end{table}

%% file: content/05-discussion.tex
\section{Discussion}
\label{sec:discussion}

\begin{figure}[h]
\centering
\includegraphics[width=1\linewidth]{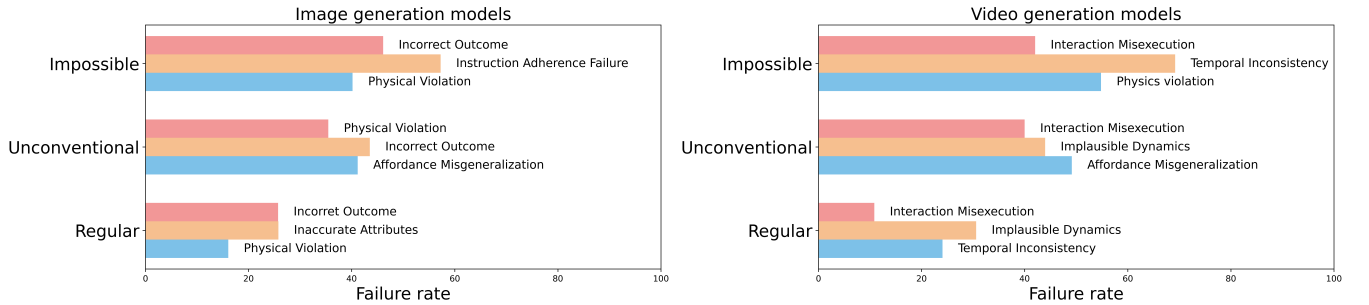}
\caption{Representative failure modes distribution across setting and scenario types.}
\label{fig:failure_modes}
\end{figure}

In this section, we provide a deeper analysis of the empirical findings revealed by \BenchName{}. Beyond quantitative performance differences, we investigate \textit{how} and \textit{why} current world models fail under long-tail physical interactions. Figure \ref{fig:failure_modes} shows the failure modes of image generation models and video generation models.

\subsection{Failure Modes Analysis of Image Generation Models}

\noindent\textbf{Regular scenarios.}
Failures are jointly led by incorrect outcomes and inaccurate attributes, followed by physical violations. Incorrect outcomes refer to cases where the expected result does not occur. Inaccurate attributes refer to objects having incorrect physical properties such as size, shape, or material. Physical violations occur when interactions break basic physical rules. For example, a hammer may be positioned correctly above a nail, but the nail is not driven in, while the hammer may also appear slightly deformed or incorrectly scaled. In some cases, objects may even slightly intersect during contact, indicating a violation of physical constraints.

\noindent\textbf{Unconventional scenarios.}
Failures are led by incorrect outcomes, followed by affordance misgeneralization and physical violations. Incorrect outcomes mean the intended effect is not achieved. Affordance misgeneralization occurs when the model recognizes relevant properties but fails to use them correctly for the task. Physical violations indicate that the interaction contradicts basic physical rules. For example, a book used to hammer a nail may be placed near the nail but fails to drive it in. The model may recognize that the book is rigid, but it does not use the flat surface to apply force, and in some cases the book may unrealistically bend during contact.

\noindent\textbf{Impossible scenarios.}
Failures are dominated by instruction adherence failure, followed by incorrect outcomes and physical violations. Instruction adherence failure occurs when the model ignores or alters the given constraints. Incorrect outcomes refer to producing a result that should not occur. Physical violations arise when interactions contradict material behavior. For example, a sponge may be shown cutting a carrot, even though it lacks the required properties. The model may ignore the constraint and implicitly treat the sponge as rigid, leading to an outcome that violates basic physical principles.

\subsection{Failure Modes Analysis of Video Generation Models} 
\label{sec:failure_modes}

\noindent\textbf{Regular scenarios.}
Failures are led by implausible dynamics, followed by interaction misexecution and temporal inconsistency. Implausible dynamics refer to unrealistic motion or lack of proper force progression. Interaction misexecution occurs when the action is attempted but not properly completed. Temporal inconsistency refers to changes in object behavior across frames. For example, a hammer may suddenly appear in contact with a nail without a visible swing, repeatedly hit the nail without driving it in, and exhibit slight jitter or discontinuity across frames.

\noindent\textbf{Unconventional scenarios.}
Failures are led by affordance misgeneralization, followed by implausible dynamics and interaction misexecution. Affordance misgeneralization means the model fails to assign the correct functional role to a tool. Implausible dynamics refer to motion or deformation that does not match physical properties. Interaction misexecution occurs when the intended action is not successfully carried out. For example, a book used to hammer a nail may initially act as a striking tool but then behave like a soft object during motion, producing unrealistic movement and failing to drive the nail in despite repeated contact.

\noindent\textbf{Impossible scenarios.}
Failures are led by temporal inconsistency, followed by physical violations and interaction misexecution. Temporal inconsistency refers to abrupt or discontinuous changes across frames. Physical violations indicate that interactions break basic physical rules. Interaction misexecution occurs when the action fails to produce a valid effect. For example, a walnut may suddenly appear cracked without any visible force buildup, while a soft tool interacts with it without deforming, and repeated contact fails to produce a physically consistent outcome.

\subsection{Sensitivity Analysis on Generation Setting}
\label{sec:sensitivity_generation}

As introduced in Section \ref{sec:benchmark}, predictive generation asks models to infer the outcome of an interaction without revealing the result, whereas descriptive generation explicitly specifies the expected outcome and requires the model to realize it visually. We further analyze the effect of generation setting to better understand the model performance and trace the error source.

\begin{figure}[t]
    \centering
    \includegraphics[width=0.32\linewidth]{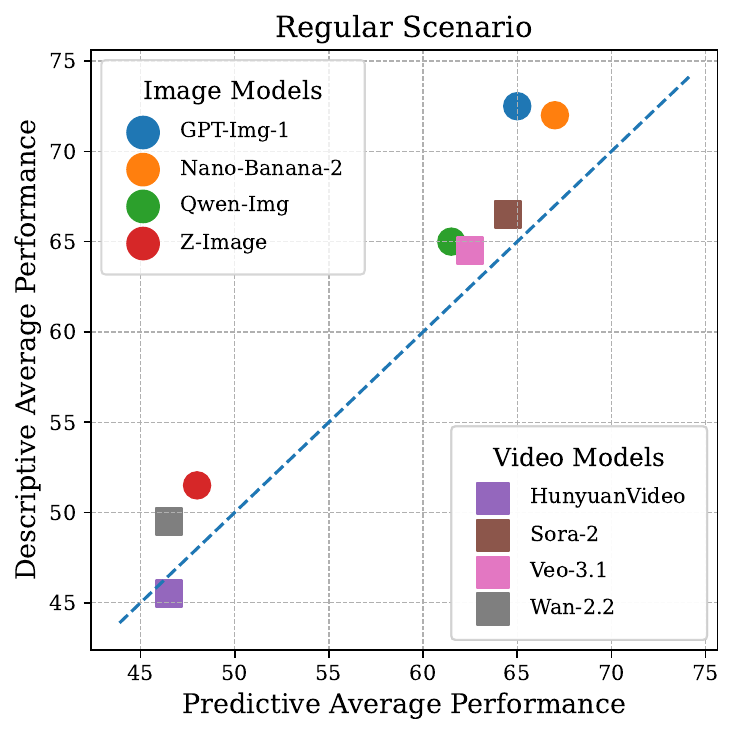}
    \includegraphics[width=0.32\linewidth]{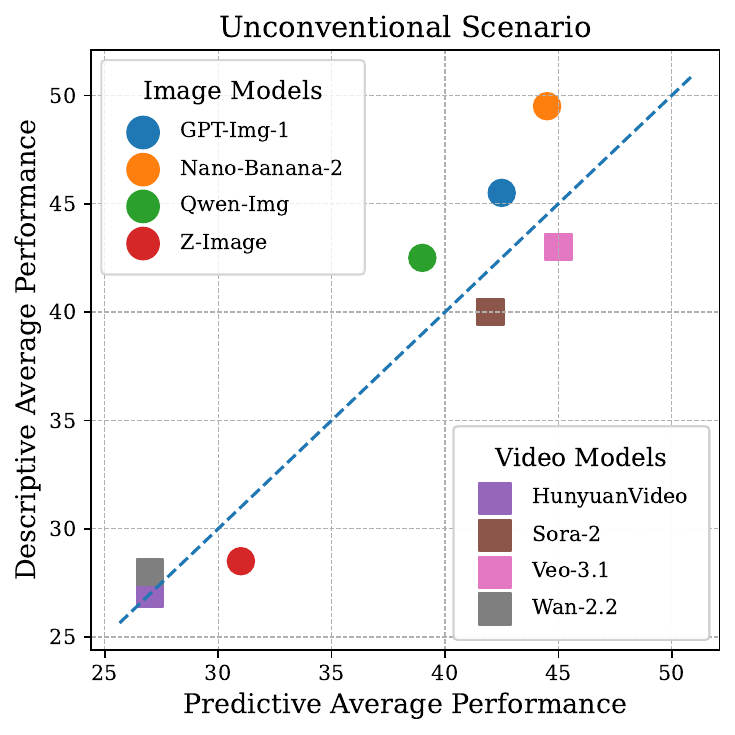}
    \includegraphics[width=0.32\linewidth]{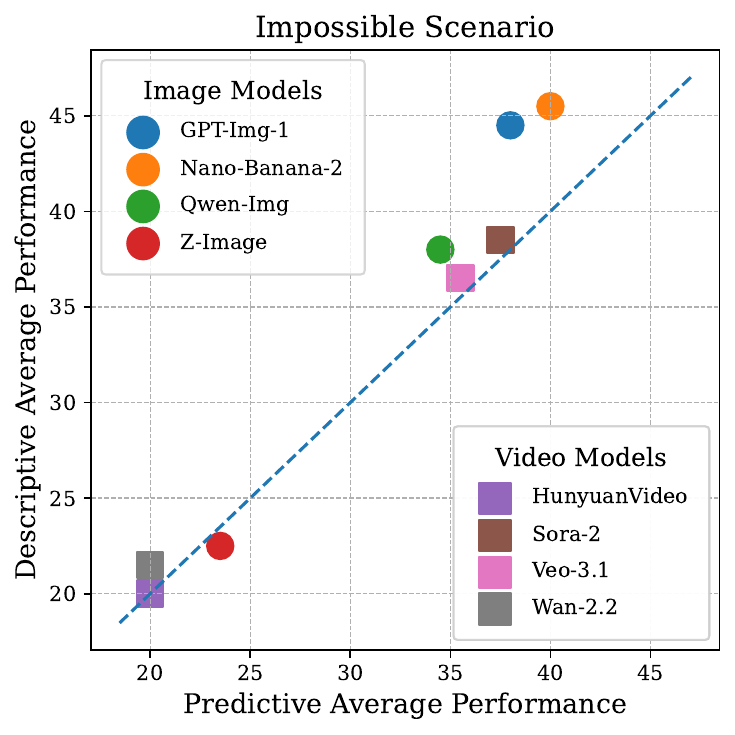}
    \caption{Predictive vs. Descriptive performance across Regular (left), Unconventional (middle), and Impossible (right) scenarios. Each point represents a model’s average score.}
    \label{fig:pred_desc_scatter}
    \vspace{-0.1in}
\end{figure}

\begin{figure}[t]
    \centering
    \includegraphics[width=1\linewidth]{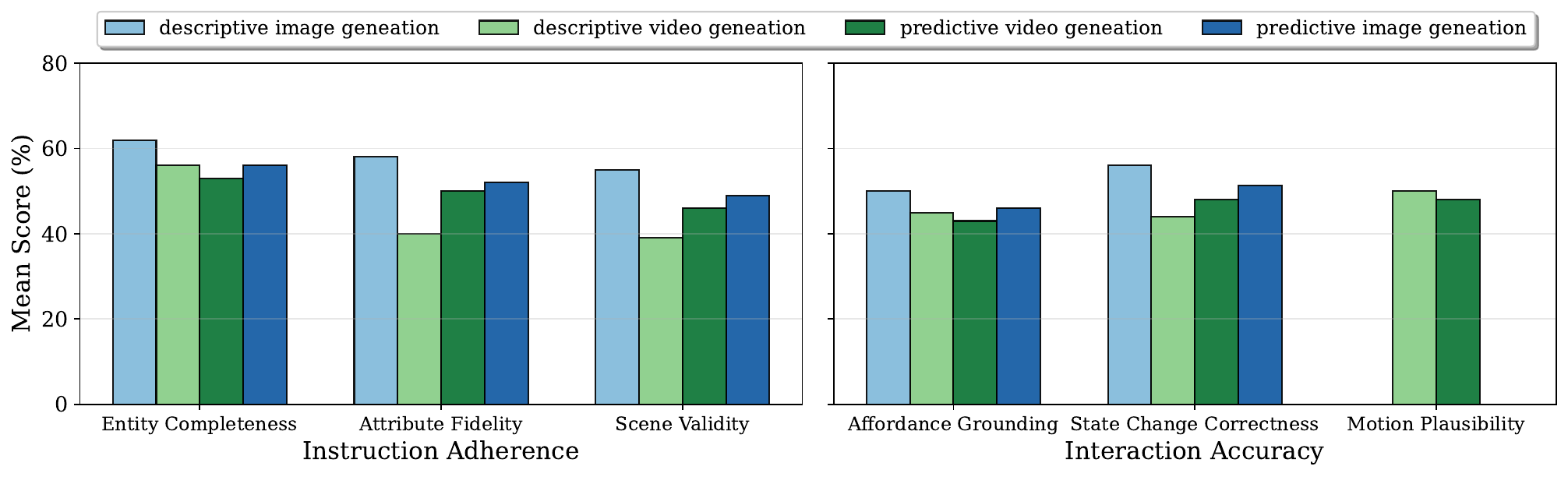}
    \caption{\textbf{Sub-metric breakdown across instruction adherence and interaction accuracy.} We report mean performance across entity completeness, attribute fidelity, scene validity, affordance grounding, state change correctness, and motion plausibility.}
    \label{fig:submetric_analysis}
    \vspace{-0.2in}
\end{figure}

Across most models, predictive generation achieves comparable or higher performance than descriptive generation, particularly for video models. As shown in Figure~\ref{fig:pred_desc_scatter}, descriptive performance often falls below predictive performance, suggesting that when explicitly instructed to produce a particular outcome, models frequently revert to familiar patterns learned from training data rather than faithfully following the prompt. The gap between the two settings becomes larger in Unconventional scenario. This indicates that when interactions deviate from common training distributions, models struggle to reconcile explicit instructions with their learned visual priors.

Figure~\ref{fig:submetric_analysis} further breaks down performance across instruction adherence and interaction accuracy metrics. Descriptive generation generally improves performance for image models, particularly on entity completeness and scene validity, suggesting that explicitly specifying the desired outcome helps guide object grounding and scene composition. In contrast, video models benefit less from descriptive prompts, with descriptive and predictive settings. Notably, attribute fidelity for video models remains relatively low with descriptive prompts. This pattern indicates that video generation models may rely more on training dataset biases and common visual patterns, leading to weaker grounding of object attributes when simulating interactions.

\subsection{Why Do World Models Struggle with Long-Tail Scenarios?}

World models fail under both \textit{Predictive generation} and \textit{Descriptive generation} settings for different reasons.

\noindent\textbf{Predictive generation.}
In the predictive setting, the model must infer the outcome of the interaction without explicit guidance. Failures in this setting suggest that the model lacks the underlying physical knowledge required for reasoning about object interactions. We hypothesize that this limitation stems from a mismatch between visual pattern learning and attribute-level physical abstraction. 

Recent image and video generation models are primarily trained to maximize perceptual realism and distributional fidelity. As a result, they tend to internalize high-frequency interaction templates (e.g., ``hammer–nail'' or ``knife–carrot'') as holistic visual patterns, rather than decomposing them into transferable physical primitives such as rigidity, sharpness, leverage, or force direction. Under \textit{Regular} scenarios, retrieving these templates is sufficient: models can generate plausible scenes by aligning objects with familiar interaction patterns. However, \textit{Unconventional} and \textit{Impossible} scenarios require compositional reasoning over physical principles (whether a tool possesses the physical properties required to accomplish the intended action). 

Such reasoning requires counterfactual evaluation (e.g., \emph{Would this object transfer sufficient force?}) rather than simple visual completion. Because attribute-level physical reasoning is only weakly supervised in current training pipelines, models often default to template interpolation or outcome completion, producing visually plausible but physically inconsistent results.

In video generation, this limitation is further amplified by temporal dynamics. While a single image may conceal local inconsistencies, multi-frame generation exposes violations of force propagation, state continuity, and physical constraints, leading to compounding errors across frames.

\noindent\textbf{Descriptive generation.}
In the descriptive setting, the expected outcome is explicitly provided in the prompt. However, models still frequently fail to generate the correct interaction. This suggests that the limitation is not only in outcome inference, but also in physically grounded simulation.

We attribute this behavior to a bias toward perceptual coherence over causal consistency. During generation, models tend to associate objects with their most common interaction patterns in the training data. As a result, even when the prompt specifies an unconventional outcome, the model often defaults to high-frequency pairings between objects and their canonical functions.

In video generation, this issue is compounded by forward propagation during sequence generation. Once an object is generated in the initial frame, the model tends to maintain its regular role across subsequent frames, continuing the familiar interaction pattern rather than following the instruction. Consequently, the model struggles to simulate the intended causal dynamics even when the expected outcome is explicitly described.

%% file: content/06-conclusion.tex
\vspace{-0.1in}
\section{Conclusion and Future Works}
\label{sec:conclusion}
\vspace{-0.1in}

In this work, we introduced \BenchName{}, a benchmark for evaluating whether visual generative models truly internalize physical principles or mainly rely on statistical regularities from training data. Our experiments reveal a clear long-tail gap in current world models. Across both image and video generation, performance consistently degrades from Regular to Unconventional and Impossible scenarios, showing that current systems struggle to generalize beyond common tool--task co-occurrences. The largest drops occur in interaction accuracy and physical realism, indicating that the main limitation is not visual quality alone, but weak affordance-level understanding and insufficient causal grounding. Moreover, video models remain substantially more brittle than image models, as they must additionally maintain temporally coherent dynamics and physically valid state transitions over time. Our further analyses suggest that these failures arise for two related reasons. In predictive generation, models often fail because they do not reliably infer outcomes from transferable physical properties such as rigidity, geometry, or force transmission. In descriptive generation, even when the desired outcome is explicitly specified, models frequently revert to familiar visual patterns instead of faithfully simulating the instructed interaction. Together, these results suggest that current world models still rely heavily on memorized interaction templates rather than compositional physical reasoning.

We hope \BenchName{} can serve as a useful testbed for future research on physically grounded world modeling. One promising direction is to incorporate stronger inductive biases for object attributes, affordances, and causal dynamics during training, so that models learn reusable physical primitives rather than holistic templates. Another direction is to improve video generation with mechanisms for long-horizon state tracking, force-consistent motion modeling, and constraint-aware temporal planning. Beyond tool-use scenarios, future benchmarks could also extend to richer embodied settings involving multi-step manipulation, multi-object causal chains, and more complex environment dynamics. We believe progress on these directions is essential for building world models that can reason about the long tail of physical interactions, rather than merely reproducing the visual statistics of common experiences.

%% file: content/supp.tex
\clearpage
\appendix
\setcounter{page}{1}

\input{supp/code_style}

\input{supp/bench_details}
\input{supp/evaluation_details}
\input{supp/more_disussions}

%% file: supp/code_style.tex
\definecolor{codegreen}{rgb}{0.0,0.5,0.0}
\definecolor{codegray}{rgb}{0.45,0.45,0.45}
\definecolor{codepurple}{rgb}{0.5,0.0,0.6}
\definecolor{backcolour}{rgb}{0.97,0.97,0.97}

\lstdefinestyle{mystyle}{
    backgroundcolor=\color{backcolour},
    basicstyle=\ttfamily\footnotesize,
    keywordstyle=\color{black},
    commentstyle=\color{codegreen}\itshape,
    stringstyle=\color{codepurple},
    breaklines=true,
    breakatwhitespace=true,
    keepspaces=true,
    showspaces=false,
    showstringspaces=false,
    showtabs=false,
    tabsize=2,
    frame=single,
    rulecolor=\color{black!0},
    captionpos=b
}
\lstset{style=mystyle}

%% file: supp/bench_details.tex
\section{Benchmark Details}
\label{app:dataset}

\subsection{Use of LLMs for Data Curation}

We choose \texttt{GPT-5} as the primary LLM for all generation steps in the data curation pipeline. The LLM is used to (1) instantiate concrete tool-use tasks from action definitions, (2) propose unconventional tool substitutions that satisfy required physical attributes, (3) generate impossible tools by reversing key affordance properties, and (4) generate evaluation prompt and rubric questions.

\subsection{Generation Prompt Templates}

We provide the full prompt templates used in each step of the data curation process described in Section 4.

\subsubsection{Step 1.Action-to-Task Generation}

The first step converts an action definition into a set of concrete, realistic, and visually demonstrable task scenarios. Each generated task includes a conventional tool, an expected outcome, and the required tool attributes derived from the action's physical affordance.

\begin{lstlisting}[language=Python,basicstyle=\ttfamily\small,breaklines=true,columns=fullflexible,keepspaces=true]
You are given an action from an action ontology.
Your task is to generate 20 diverse, realistic task scenarios that require this action.
Each task must include:
1. A specific task goal (what needs to be accomplished, please make it as common as possible)
2. An original/conventional tool that would be used for this task
3. The expected outcome of completing the task
4. Required tool attributes that enable the action (based on physics and affordance)

Action Information:
- Action Name: {action_name}
- Action Description: {action_description}
- Physics: {physics}
- Affordance: {affordance}

Task Scenario Requirements:
Each task scenario should be simple, direct, and visually clear enough that the entire process from start to successful outcome could be convincingly demonstrated in a 5 second video clip. Focus on actions that can be completed swiftly, show obvious visible change, and do not require prolonged, hidden, or ambiguous steps. Ensure that for every task, an observer could identify the action, tool, and result within a brief visual sequence.
Generate exactly 20 tasks. Be CREATIVE and DIVERSE across scenarios - vary contexts (home, kitchen, workshop, office, outdoor, etc.), materials, scales, and use cases. Avoid repetition. At the same time, keep each task REALISTIC: specific, concrete, and plausible in everyday life. Each task should clearly require the given action.

Output your response as a JSON object with the following structure:
{
    "tasks": [
        {
            "task_goal": "a specific task description (e.g., 'tighten a loose screw on a chair')",
            "original_tool": "the conventional tool name (e.g., 'screwdriver')",
            "expected_outcome": "what happens when the task is completed successfully",
            "required_tool_attributes": [
                "attribute 1 (e.g., 'narrow tip')",
                "attribute 2 (e.g., 'rigid structure')",
                "attribute 3 (e.g., 'torque transmission')"
            ]
        },
        ... (20 tasks total)
    ]
}

The required_tool_attributes for each task should be derived from the physics and affordance information. Think about what physical properties the tool must have to enable the action.

Return ONLY the JSON object, no additional text.
\end{lstlisting}

\subsubsection{Step 2: Unconventional Tool Generation}

Given a task scenario and its required attributes, the second Step generates
unconventional but physically plausible substitute tools. These tools are not
the canonical choice, but they may still work because they share relevant
functional properties with the original tool.

\begin{lstlisting}[language=Python,basicstyle=\ttfamily\small,breaklines=true,columns=fullflexible,keepspaces=true]
You are given a task scenario with its original tool and required attributes. Your task is to identify unconventional tools that could potentially accomplish the task (tools not typically used but might work).

Task Information:
- Task Goal: {task_goal}
- Original Tool: {original_tool}
- Expected Outcome: {expected_outcome}
- Required Tool Attributes: {required_tool_attributes}

For unconventional_tools:
- Think of everyday objects that have some of the required attributes but aren't the conventional choice
- These should be objects that might work but are less ideal (e.g., using a coin instead of a screwdriver)
- Include 4-5 examples that match the required attributes

Object-Affordance Graph for reference: 
{OAG}

Output your response as a JSON object with the following structure:
{
    "unconventional_tools": [
        "tool 1",
        "tool 2"
    ]
}

Return ONLY the JSON object, no additional text.
\end{lstlisting}

\subsubsection{Step 3: Opposite Attribute and Impossible Tool Generation}

The third Step constructs physically incompatible tools by first identifying
attributes that oppose the required functional properties and then generating
objects that clearly cannot complete the task. These instances are used to test
constraint awareness and failure recognition.

\begin{lstlisting}[language=Python,basicstyle=\ttfamily\small,breaklines=true,columns=fullflexible,keepspaces=true]
You are given a task scenario with its required attributes. Your task is to:
1. Identify opposite tool attributes (attributes that would make the task difficult or impossible)
2. Identify tools that would be IMPOSSIBLE to use for this task

Task Information:
- Task Goal: {task_goal}
- Original Tool: {original_tool}
- Required Tool Attributes: {required_tool_attributes}

For opposite_tool_attributes:
- These are attributes that directly oppose or conflict with the required attributes
- Think about what would make the tool ineffective (e.g., if "rigid structure" is required, "soft material" would be opposite)
- Include 2-3 examples

For impossible_tools:
- Generate a list of tools/objects that would be impossible to use for this task because they:
  - Have the opposite attributes (e.g., soft, flexible, rounded when rigid, sharp is needed)
  - Lack any of the critical required attributes
  - Are fundamentally incompatible with the physics of the action
- These should be objects that clearly cannot accomplish the task.
- Include 4-5 examples

Object-Affordance Graph for reference: 
{OAG}

Output your response as a JSON object with the following structure:
{
    "opposite_tool_attributes": [
        "opposite attribute 1",
        "opposite attribute 2"
    ],
    "impossible_tools": [
        "tool 1",
        "tool 2",
        "tool 3",
        "tool 4"
    ]
}

Return ONLY the JSON object, no additional text.
\end{lstlisting}

\subsubsection{Step 4: Evaluation Instance and Prompt Construction}

Each task is expanded into five evaluation instances: one regular tool, two
unconventional tools, and two impossible tools. For each instance, we generate
four prompts to evaluate predictive and descriptive capabilities in both image
and video generation settings.

\begin{lstlisting}[language=Python,basicstyle=\ttfamily\small,breaklines=true,columns=fullflexible,keepspaces=true]
You are helping design evaluation prompts for a physical tool-use benchmark.

For each task, you must generate FIVE evaluation instances that probe different
capabilities while remaining physically realistic and visually plausible.

For this task, you are given:
- task_goal: {task_goal}
- original_tool: {original_tool}
- expected_outcome (for successful completion with a suitable tool):
  {expected_outcome}
- unconventional_tools: {unconventional_tools}
- impossible_tools: {impossible_tools}

Your job is to create FIVE evaluation instances:
1. One instance using the original_tool (tool_type = "regular").
2. Two instances using distinct unconventional_tools (tool_type = "unconventional").
3. Two instances using distinct impossible_tools (tool_type = "impossible").

Each evaluation instance must contain FOUR prompts:

(a) Predictive IMAGE prompt.
- An image-generation prompt that asks model to generate the outcome state of applying tool X to object Y for task Z, without revealing the final result.
- Asks what will happen when applying tool X to object/scenario Y for task Z,
  WITHOUT revealing the final outcome in the prompt text.
- It should sound like a natural user query aimed at generating a single image
  depicting the anticipated situation, but must not state success or failure.
- One example: "Generate an image to visualize the final state of using the book to hammer the nail. Be realistic"

(b) Descriptive IMAGE prompt.
- An image-generation prompt that asks model to generate the outcome state of applying tool X to object Y for task Z by describing the final state.
- For SUCCESSFUL (regular/unconventional) tools:
  - Describe ONLY the final visible state after the task has been completed
    successfully, consistent with `expected_outcome`.
- For IMPOSSIBLE tools:
  - Describe ONLY the final visible failure state (what things look like after the
    failed attempt), inferred from physics and tool limitations.
- Focus on the static final configuration (no process description, no multi-step wording).

(c) Predictive VIDEO prompt.
- A video-generation prompt that asks model to predict and simulate asking it to anticipate the outcome of applying tool X to object Y for task Z, without revealing the final result.
- Asks what will happen when a person uses the tool for the task in a short video,
  WITHOUT revealing whether the attempt succeeds or fails.
- Describe the setup and intended interaction in a way that suggests a short clip,
  but do not say if the outcome is success or failure.
- One example: "Generate a video to illustrate what will happen when using the book to hammer the nail? Be realistic. The video should contain the full process from the start state to the end state."

(d) Descriptive VIDEO prompt.
- A video-generation prompt for a short video showing the full process and the final state.
- For SUCCESSFUL (regular/unconventional) tools:
  - Describe how a person uses the tool on the object over time and end with the
    successful final state consistent with `expected_outcome`.
- For IMPOSSIBLE tools:
  - Describe how the person attempts the task, how and why it fails, and end with
    the visible failure state.

CRITICAL REQUIREMENTS:
- Use natural, fluent English without mentioning "task_goal", "original_tool",
  "unconventional", or "impossible" explicitly inside the prompts.
- Do NOT reveal labels like "regular", "unconventional", or "impossible" inside
  the prompts; those are only for the JSON metadata.
- For predictive prompts:
  - Never state the final outcome explicitly (success or failure).
- For descriptive prompts:
  - Always make the intended outcome (success for regular/unconventional tools,
    failure for impossible tools) explicit and visually checkable.
- Ensure that each of the FIVE instances corresponds to a single specific tool.
- Prefer concise prompts that could realistically be input by a user.

OUTPUT FORMAT (IMPORTANT):
Return ONLY a JSON object with the following structure and no extra text:
{
  "evaluation_instances": [
    {
      "tool_type": "regular" | "unconventional" | "impossible",
      "tool": "the specific tool name you used for this instance",
      "expected_outcome": "the description of the expected outcome for successful tools, the description of the failure state for impossible tools",
      "predictive_image_prompt": "a single natural-language predictive IMAGE-generation prompt",
      "descriptive_image_prompt": "a single natural-language descriptive IMAGE-generation prompt for the final state",
      "predictive_video_prompt": "a single natural-language predictive VIDEO-generation prompt",
      "descriptive_video_prompt": "a single natural-language descriptive VIDEO-generation prompt for the process and final state",
    },
    ... (exactly five instances total)
  ]
}

- Ensure there are EXACTLY five entries in "evaluation_instances":
  - 1 with tool_type = "regular"
  - 2 with tool_type = "unconventional"
  - 2 with tool_type = "impossible"
- Ensure that the `tool` for each instance appears in the appropriate list:
  - regular: original_tool
  - unconventional: from unconventional_tools
  - impossible: from impossible_tools
\end{lstlisting}

\subsubsection{Step 5: Rubric Generation}

For each evaluation instance, we generate four structured rubrics corresponding to predictive image, descriptive image, predictive video, and descriptive video outputs. Each rubric measures both instruction adherence and interaction accuracy through fine-grained checklist items.

\begin{lstlisting}[language=Python,basicstyle=\ttfamily\small,breaklines=true,columns=fullflexible,keepspaces=true]
You are designing checklist-based evaluation RUBRICS for a SINGLE
evaluation instance in a physical tool-use benchmark. For this one instance,
you must produce FOUR separate rubrics:
- one for the predictive IMAGE prompt,
- one for the descriptive IMAGE prompt,
- one for the predictive VIDEO prompt,
- one for the descriptive VIDEO prompt.

You are given:
- task_goal: {task_goal}
- action_type: {action_type}
- original_tool: {original_tool}
- expected_outcome_for_task: {expected_outcome}
- required_tool_attributes: {required_tool_attributes}
- unconventional_tools: {unconventional_tools}
- impossible_tools: {impossible_tools}

And for ONE specific evaluation instance you are also given:
- tool_type: {tool_type}   # "regular" | "unconventional" | "impossible"
- tool_name: {tool}
- instance_expected_outcome: {instance_expected_outcome}
- predictive_image_prompt: {predictive_image_prompt}
- descriptive_image_prompt: {descriptive_image_prompt}
- predictive_video_prompt: {predictive_video_prompt}
- descriptive_video_prompt: {descriptive_video_prompt}

Your job is to generate FOUR structured, checklist-based RUBRICS that can
later be used to score model-generated outputs for THIS instance:
- predictive_image_rubric   -> for predictive_image_prompt (images only)
- descriptive_image_rubric  -> for descriptive_image_prompt (images only)
- predictive_video_rubric   -> for predictive_video_prompt (videos only)
- descriptive_video_rubric  -> for descriptive_video_prompt (videos only)

The rubric has TWO top-level dimensions, each with THREE sub-dimensions:

1) Instruction Adherence (0-100%)
   Measures whether the generated scene correctly instantiates the entities and
   functional properties required to enable the intended interaction.

   It is decomposed into three sub-dimensions:

   (a) Entity Completeness
       The presence of all required entities.
       Example questions:
       - Is the specified tool present?
       - Is the target object instantiated?
       - Are required contextual elements (supporting surface, environment) included?

   (b) Attribute Fidelity
       Correct instantiation of required functional attributes.
       Example questions:
       - Does the tool exhibit the required functional property
         (e.g., rigidity or sharpness)?
       - Is the material consistent with intended physical behavior?
       - Are size and structural properties compatible with the task?

   (c) Scene Validity
       Spatial configuration and physically feasible arrangement.
       Example questions:
       - Is the tool positioned at the correct interaction region?
       - Is the relative scale between tool and object plausible?
       - Does the arrangement enable the intended interaction?

2) Interaction Accuracy (0-100%)
   Measures whether the interaction outcome and dynamics are correctly realized.

   It is decomposed into three sub-dimensions:

   (a) State Change Correctness
       Correctness of the physically correct final state (or predicted outcome).
       Example questions:
       - Does the object exhibit the correct resulting state
         (e.g., cracked, cut, bent)?
       - In impossible cases, is failure correctly depicted?
       - Is the final configuration consistent with the applied interaction?

   (b) Affordance Grounding
       Whether interaction behavior aligns with object affordances.
       Example questions:
       - Is force applied through a structurally appropriate part of the tool?
       - Is the interaction consistent with object geometry and material constraints?
       - Does the behavior reflect a physically plausible affordance?

   (c) Motion Plausibility  (ONLY for video generation)
       Temporal coherence and dynamical feasibility.
       Example questions:
       - Is motion trajectory continuous?
       - Are deformations temporally consistent?

For EACH of the four rubrics, you must create checklist questions that are:
- SPECIFIC to this task_goal, tool, and evaluation instance;
- Grounded in the provided prompts and expected outcomes;
- Physically meaningful and visually checkable.

Do NOT score anything yourself. Only define the questions.

OUTPUT FORMAT (IMPORTANT)

Return ONLY a JSON object with this structure and no extra text:

{
  "predictive_image_rubric": {
    "instruction_adherence": {
      "entity_completeness": {
        "checklist_items": [
          {
            "id": "short_snake_case_identifier",
            "question": "clear yes/no style question about entity completeness for the predictive IMAGE output"
          }
        ]
      },
      "attribute_fidelity": {
        "checklist_items": [
          {
            "id": "short_snake_case_identifier",
            "question": "clear yes/no style question about attributes for the predictive IMAGE output"
          }
        ]
      },
      "scene_validity": {
        "checklist_items": [
          {
            "id": "short_snake_case_identifier",
            "question": "clear yes/no style question about spatial / scene validity for the predictive IMAGE output"
          }
        ]
      }
    },
    "interaction_accuracy": {
      "state_change_correctness": {
        "checklist_items": [
          {
            "id": "short_snake_case_identifier",
            "question": "clear yes/no style question about final state correctness for the predictive IMAGE output"
          }
        ]
      },
      "affordance_grounding": {
        "checklist_items": [
          {
            "id": "short_snake_case_identifier",
            "question": "clear yes/no style question about affordance-consistent usage for the predictive IMAGE output"
          }
        ]
      },
      "motion_plausibility": {
        "checklist_items": [
        ]
      }
    }
  },
  "descriptive_image_rubric": {
    "instruction_adherence": {
      "entity_completeness": {
        "checklist_items": [
          {
            "id": "short_snake_case_identifier",
            "question": "clear yes/no style question about entity completeness for the descriptive IMAGE output"
          }
        ]
      },
      "attribute_fidelity": {
        "checklist_items": [
          {
            "id": "short_snake_case_identifier",
            "question": "clear yes/no style question about attributes for the descriptive IMAGE output"
          }
        ]
      },
      "scene_validity": {
        "checklist_items": [
          {
            "id": "short_snake_case_identifier",
            "question": "clear yes/no style question about spatial / scene validity for the descriptive IMAGE output"
          }
        ]
      }
    },
    "interaction_accuracy": {
      "state_change_correctness": {
        "checklist_items": [
          {
            "id": "short_snake_case_identifier",
            "question": "clear yes/no style question about final state correctness for the descriptive IMAGE output"
          }
        ]
      },
      "affordance_grounding": {
        "checklist_items": [
          {
            "id": "short_snake_case_identifier",
            "question": "clear yes/no style question about affordance-consistent usage for the descriptive IMAGE output"
          }
        ]
      },
      "motion_plausibility": {
        "checklist_items": [
        ]
      }
    }
  },
  "predictive_video_rubric": {
    "instruction_adherence": {
      "entity_completeness": {
        "checklist_items": [
          {
            "id": "short_snake_case_identifier",
            "question": "clear yes/no style question about entity completeness for the predictive VIDEO output"
          }
        ]
      },
      "attribute_fidelity": {
        "checklist_items": [
          {
            "id": "short_snake_case_identifier",
            "question": "clear yes/no style question about attributes for the predictive VIDEO output"
          }
        ]
      },
      "scene_validity": {
        "checklist_items": [
          {
            "id": "short_snake_case_identifier",
            "question": "clear yes/no style question about spatial / scene validity for the predictive VIDEO output"
          }
        ]
      }
    },
    "interaction_accuracy": {
      "state_change_correctness": {
        "checklist_items": [
          {
            "id": "short_snake_case_identifier",
            "question": "clear yes/no style question about final state correctness for the predictive VIDEO output"
          }
        ]
      },
      "affordance_grounding": {
        "checklist_items": [
          {
            "id": "short_snake_case_identifier",
            "question": "clear yes/no style question about affordance-consistent usage for the predictive VIDEO output"
          }
        ]
      },
      "motion_plausibility": {
        "checklist_items": [
          {
            "id": "short_snake_case_identifier",
            "question": "clear yes/no style question about temporal / motion plausibility for the predictive VIDEO output"
          }
        ]
      }
    }
  },
  "descriptive_video_rubric": {
    "instruction_adherence": {
      "entity_completeness": {
        "checklist_items": [
          {
            "id": "short_snake_case_identifier",
            "question": "clear yes/no style question about entity completeness for the descriptive VIDEO output"
          }
        ]
      },
      "attribute_fidelity": {
        "checklist_items": [
          {
            "id": "short_snake_case_identifier",
            "question": "clear yes/no style question about attributes for the descriptive VIDEO output"
          }
        ]
      },
      "scene_validity": {
        "checklist_items": [
          {
            "id": "short_snake_case_identifier",
            "question": "clear yes/no style question about spatial / scene validity for the descriptive VIDEO output"
          }
        ]
      }
    },
    "interaction_accuracy": {
      "state_change_correctness": {
        "checklist_items": [
          {
            "id": "short_snake_case_identifier",
            "question": "clear yes/no style question about final state correctness for the descriptive VIDEO output"
          }
        ]
      },
      "affordance_grounding": {
        "checklist_items": [
          {
            "id": "short_snake_case_identifier",
            "question": "clear yes/no style question about affordance-consistent usage for the descriptive VIDEO output"
          }
        ]
      },
      "motion_plausibility": {
        "checklist_items": [
          {
            "id": "short_snake_case_identifier",
            "question": "clear yes/no style question about temporal / motion plausibility for the descriptive VIDEO output"
          }
        ]
      }
    }
  }
}

DETAILED INSTRUCTIONS:
- The top-level JSON object MUST contain ALL FOUR keys:
  - "predictive_image_rubric"
  - "descriptive_image_rubric"
  - "predictive_video_rubric"
  - "descriptive_video_rubric"
- Each of those four rubric objects MUST have the SAME internal structure:
  - "instruction_adherence" with sub-dimensions "entity_completeness",
    "attribute_fidelity", "scene_validity", each with a "checklist_items" list.
  - "interaction_accuracy" with sub-dimensions "state_change_correctness",
    "affordance_grounding", "motion_plausibility", each with a "checklist_items" list
    (for image rubrics, "motion_plausibility.checklist_items" may be empty but must exist).
- For EVERY sub-dimension in EVERY rubric:
  - Use 3-6 checklist items when they are meaningful, except for cases where
    motion is not applicable to images (then you may use 0-2 highly specific items).
  - Each checklist item MUST include:
    - "id": a short, unique snake_case identifier (no spaces, no punctuation),
    - "question": a concise, self-contained question that can be judged from
      the generated media for that specific prompt type.
- Tailor the wording of each question to THIS specific task, tool, instance, and
  prompt type (predictive vs descriptive, image vs video).
- Reflect whether the instance is regular / unconventional / impossible when
  phrasing questions about success vs. failure states.
- Do NOT mention internal labels like "instruction adherence" or
  "interaction accuracy" inside the question text itself; those are implicit in
  the JSON structure.
- Do NOT include any comments or explanations outside the JSON. Only return
  the JSON object described above.
\end{lstlisting}

\subsection{Human Verification}

To ensure dataset quality, we conduct two rounds of manual verification. Four human volunteers participate in the annotation process, including three master's students in computer science and one undergraduate student majoring in physics. 

In the first round, annotators review the automatically generated task candidates and remove ambiguous, unrealistic, or visually unclear examples. They also verify whether the proposed unconventional tools satisfy the required functional attributes and whether the impossible tools clearly violate the critical physical constraints of the task. 

In the second round, annotators review the generated prompts and evaluation rubrics to correct unclear descriptions and ensure consistency between task specifications and evaluation criteria.

 Specifically, annotators examine: (1) whether each task is realistic and visually demonstrable, (2) whether unconventional tools are physically plausible substitutes given the required attributes, (3) whether impossible tools genuinely violate the physical or affordance constraints of the task, (4) whether predictive and descriptive prompts are clear and visually verifiable, and (5) whether rubric checklist items are specific, observable, and aligned with the intended interaction outcome.

%% file: supp/evaluation_details.tex
\section{Evaluation Details}
\label{app:eval}

We provide additional details on the evaluation protocol of \BenchName{}. 

\subsection{Evaluation Data}

Due to limited annotation and computational resources, we sample a subset of prompts from the full benchmark for human evaluation. Specifically, we select 80 high-quality prompts covering different scenario modes (Regular, Unconventional, and Impossible) and generation settings (Predictive and Descriptive). These prompts are used to generate outputs for each evaluated model. We collected 1080 generated data points for evaluation.

Note that some prompts cannot be generated by certain models (e.g., Sora-2 and Veo3), likely due to safety filtering policies. In such cases, we replace the blocked prompts with alternative prompts sampled from the full benchmark to maintain the target number of evaluation instances.

\subsection{Human Annotation}

To ensure reliable assessment of generated outputs, we conduct human evaluation with annotators who have prior experience in visual content evaluation. 

\noindent\textbf{Annotator Recruitment.}
We recruit nine annotators with backgrounds in computer vision and multimodal generation. All annotators are computer science students, including four PhD students, three undergraduate students, and two master’s students. Each annotator receives a detailed guideline document explaining the benchmark objectives and evaluation criteria. All annotators participated on a voluntary basis, and we sincerely thank them for their contributions to the annotation process.

\begin{figure}[h]
    \centering
    \includegraphics[width=1\linewidth]{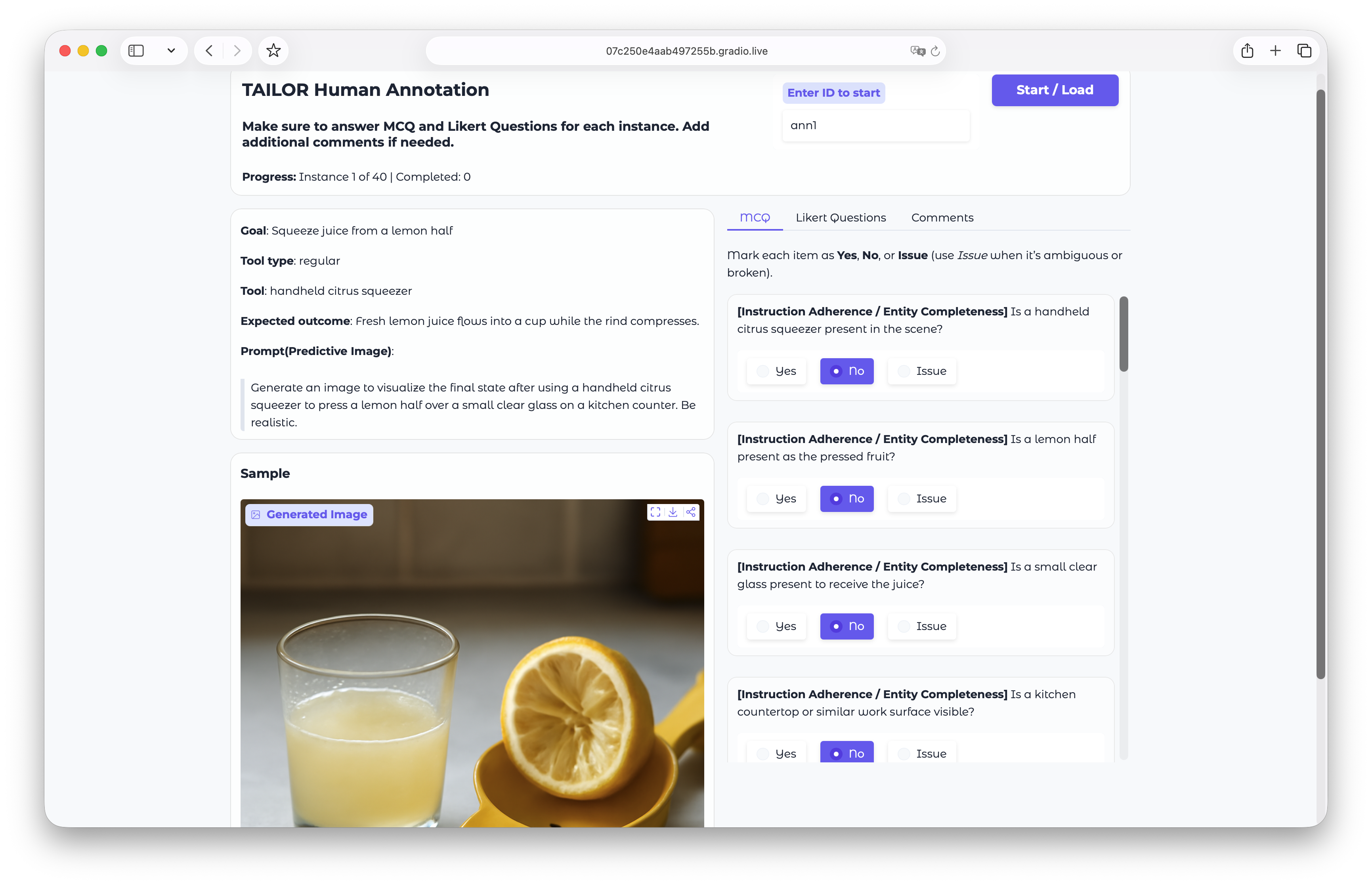}
    \caption{Human Annotation Interface for Rubric Questions}
    \label{fig:ui_1}
\end{figure}

\begin{figure}[h]
    \centering
    \includegraphics[width=1\linewidth]{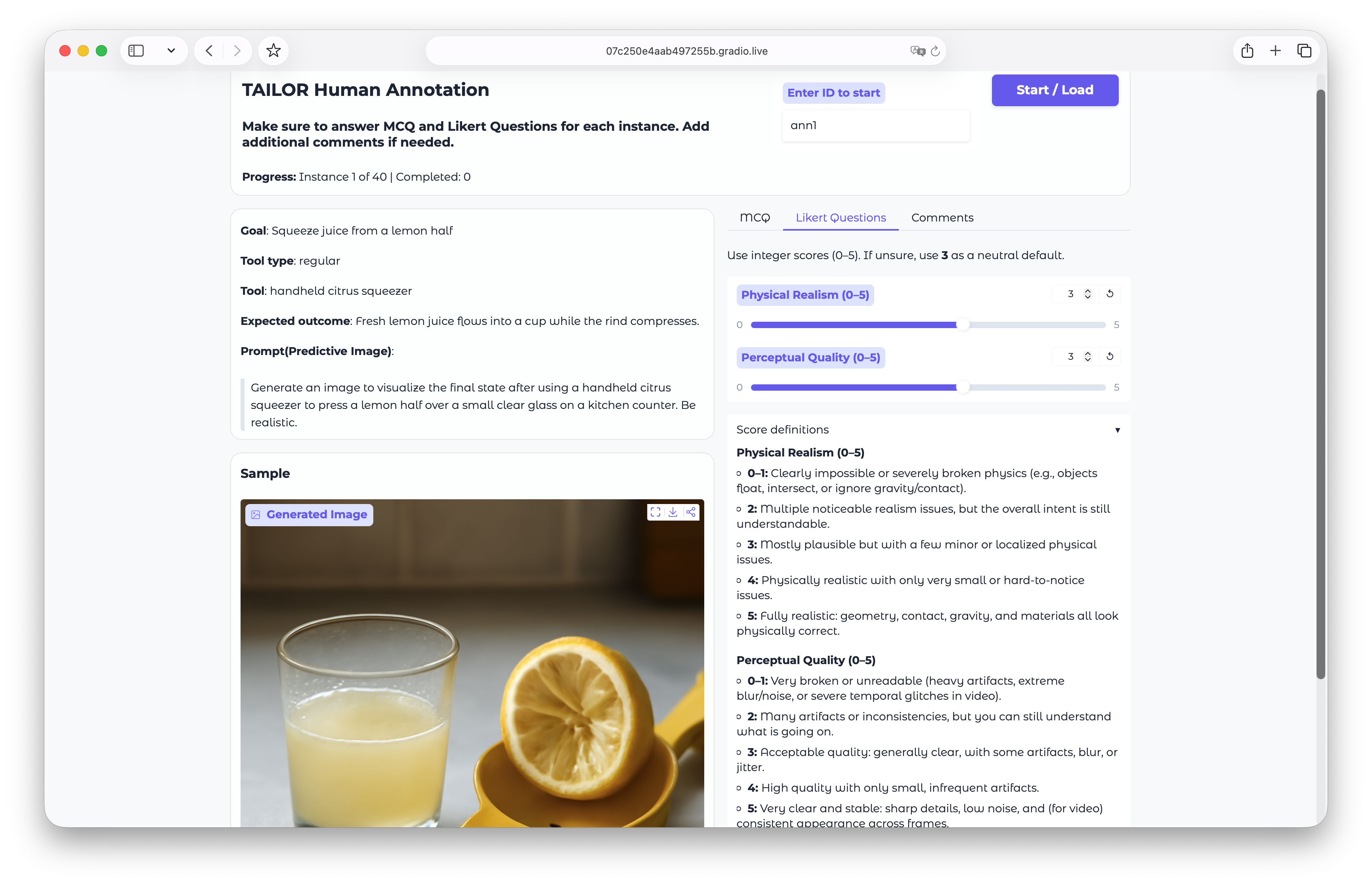}
    \caption{Human Annotation Interface for Open-ended Rating Questions}
    \label{fig:ui_2}
\end{figure}

\noindent\textbf{Evaluation Interface.}
Each generated sample is presented together with the prompt and the corresponding evaluation questions. Annotators evaluate the sample by answering a set of checklist-based questions (see Figure \ref{fig:ui_1}) and rating open-ended quality questions (see Figure \ref{fig:ui_2}).

\noindent\textbf{Annotation Procedure.}
For each generated sample, annotators perform the following steps:

\begin{enumerate}
\item Read the prompt describing the interaction scenario.
\item Observe the generated image or video.
\item Answer checklist questions corresponding to the structured metrics.
\item Assign scores for open-ended quality metrics.
\item (Optional) Add comments for current sample or evaluation questions.
\end{enumerate}

Each sample is independently evaluated by at least three annotators. The final human score is obtained by averaging the ratings across annotators.

\noindent\textbf{Inter-Annotator Agreement.}
We measure inter-annotator agreement across human evaluators. For the checklist-based metrics, we compute Krippendorff’s $\alpha$ to quantify agreement among multiple annotators. We obtain an average $\alpha$ of 0.72, indicating substantial agreement. For the open-ended quality scores, we report the average pairwise Spearman correlation between annotators, which is 0.68, further confirming consistent scoring behavior.

\subsection{Automatic Evaluation with VLM Judge}

In addition to human evaluation, we employ a vision-language model (VLM) 
as an automatic judge to improve scalability and reproducibility.

\noindent\textbf{Judge Model.}
We use \texttt{gemini-2.5-pro} as the automatic evaluation model. The judge 
is provided with the same information available to human annotators, 
including the prompt, the generated media, and the evaluation rubric.

\noindent\textbf{Evaluation Prompt.}
For each sample, the judge receives instructions to answer the rubric questions and produce structured scores for each metric. To improve  reliability, the judge is prompted to first explain its reasoning and  then produce the final score.

\begin{lstlisting}
PROMPT = """
You are an expert evaluator for visual physical interaction tasks.

Your role is to assess whether a generated image or video correctly depicts a physical interaction described in the prompt. The evaluation focuses on object presence, physical attributes, interaction behavior, and the resulting
state change. Carefully analyze the generated sample and answer the rubric questions to produce structured scores for each metric.

Evaluation Criteria:
{questions}

Evaluation Guidelines:
- Base your judgment only on what is visible in the generated sample.
- Do not assume missing objects, attributes, or actions.
- If a required element is absent or unclear, reduce the corresponding score.
- Be strict about physical plausibility and object affordances.
- Ensure all scores are consistent with the visual evidence.
- Keep scoring consistent across different samples.

Output Format:
{
    "q1": <answer>,
    "q2": <answer>,
    ...
}
"""
\end{lstlisting}
